%% file: arxiv_version.tex
\documentclass[10pt,twocolumn,letterpaper]{article}

\PassOptionsToPackage{table}{xcolor}
\usepackage[pagenumbers]{cvpr} 

\input{preamble}

\usepackage{float}
\usepackage{tabularx}
\usepackage{array}
\usepackage{multicol}
\usepackage{multirow}
\usepackage{makecell}
\usepackage{needspace}
\usepackage{algorithmic}
\usepackage{algorithm}
\usepackage{wrapfig}

\usepackage{titletoc}

\definecolor{gaincolor}{HTML}{2E7D32}
\definecolor{losscolor}{HTML}{C62828}

\newcommand{\TaskName}[1]{\textit{#1}}
\newcommand{\phasearrow}{\ensuremath{\longrightarrow}}
\newcommand{\actionstandard}[5]{%
  \par\addvspace{0.3em}%
  \noindent
  \begin{minipage}{\linewidth}
    \subsection{#1}

    \noindent\textbf{Reference task:} \TaskName{#3}\par
    \vspace{0.1em}
    \noindent\textbf{Definition:} #2\par

    \vspace{0.25em}
    \noindent\makebox[\linewidth][c]{%
      \begin{minipage}[c]{0.20\linewidth}
        \centering
        \includegraphics[width=\linewidth]{#4}\par
        \vspace{0.1em}
        {\footnotesize Start frame\par}
      \end{minipage}%
      \hspace{1em}%
      \phasearrow
      \hspace{1em}%
      \begin{minipage}[c]{0.20\linewidth}
        \centering
        \includegraphics[width=\linewidth]{#5}\par
        \vspace{0.1em}
        {\footnotesize End frame\par}
      \end{minipage}%
    }\par
  \end{minipage}\par
}

\definecolor{cvprblue}{rgb}{0.21,0.49,0.74}
\usepackage[pagebackref,breaklinks,colorlinks,allcolors=cvprblue]{hyperref}

\def\confName{CVPR}
\def\confYear{2027}

\title{LexiconVLA: Learning Reusable Atomic Action Codebooks for Manipulation}

\author{
Zeming Wei$^{1}$\thanks{Equal contribution.} \quad
Jianheng Ye$^{1}$\footnotemark[1] \quad
Xinshuai Song$^{1}$ \quad
Sirui Chen$^{1}$ \quad
Yang Liu$^{1,2}$\thanks{Corresponding author.}  \\[0.5em]
$^{1}$Sun Yat-sen University \quad
$^{2}$X-Era AI Lab \\[0.5em]
{\tt\small \{weizm6,yejh57,songxsh,chensr63\}@mail2.sysu.edu.cn} \\
{\tt\small liuy856@mail.sysu.edu.cn}
}

\begin{document}

\twocolumn[{
  \renewcommand{\twocolumn}[1][]{#1}%
  \maketitle

  \begingroup
    \centering
    \includegraphics[width=0.96\textwidth]{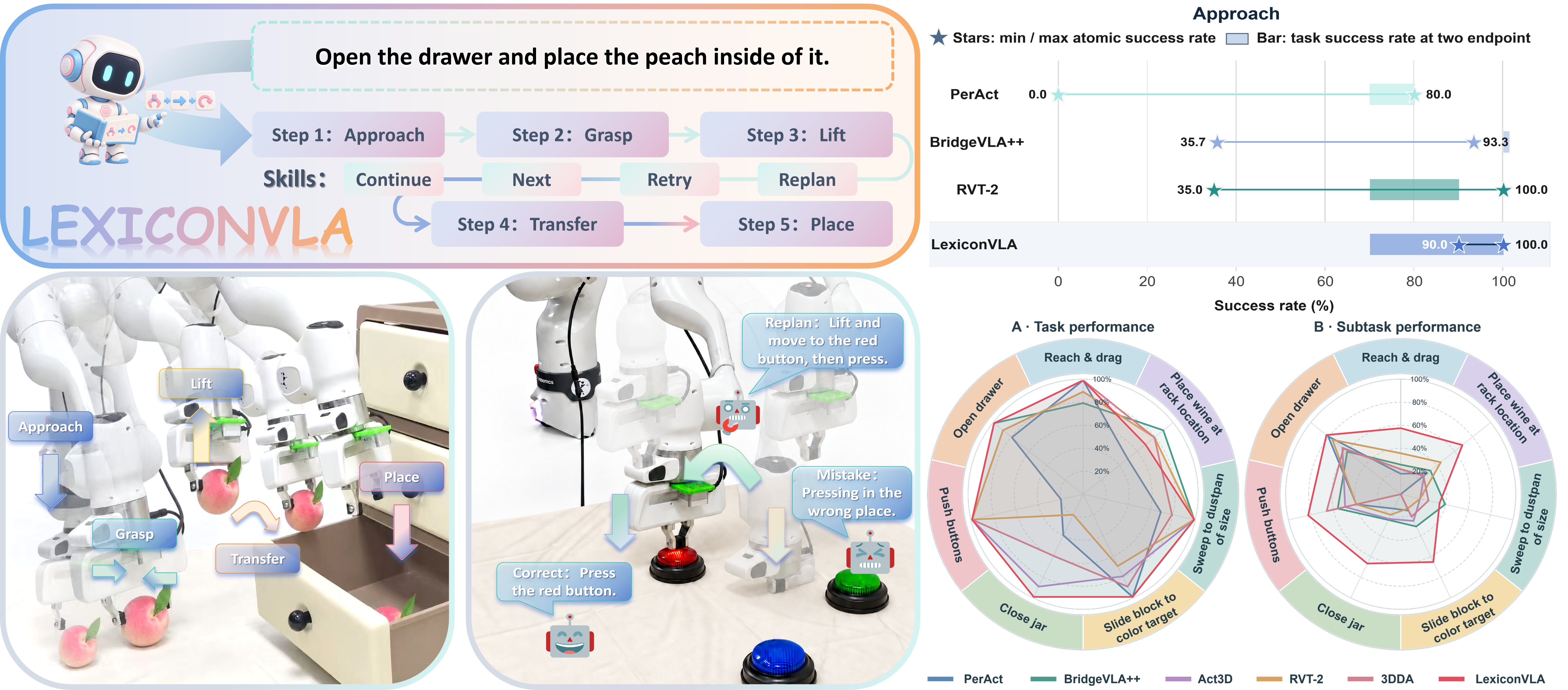}
    \captionof{figure}{\textbf{LexiconVLA and Motivation.} \textbf{Left:} LexiconVLA decomposes an instruction into atomic-action subtasks, executes each with a VLA policy conditioned on codes from a shared codebook, and advances, retries, or replans according to execution feedback; real-robot rollouts show step-wise execution and recovery from a wrong button press~(Bottom left). \textbf{Right:} A diagnostic probe on seven tasks, where each atomic subtask is prompted alone from the demonstration state. \textbf{Top right:} Lowest and highest cross-task success of the contact-free action - \textit{approach} (stars), with full-task success on the two corresponding tasks (bars). \textbf{Bottom right:} Full-task success (A) versus mean atomic-subtask success (B). For existing policies, high full-task success does not translate into reliable or task-consistent atomic execution.}
    \label{fig:Teaser}
    \par
  \endgroup
  \vspace{10pt} 
}]

\input{sec/0_Abstract}
\input{sec/1_Intro}

\input{sec/2_Related_work}

\input{sec/3_Dataset}
\input{sec/4_Method}

\input{sec/5_Experiments}
\input{sec/6_Conclusion}

{
    \small
    \bibliographystyle{ieeenat_fullname}
    \bibliography{main}
}

\input{sec/X_suppl}

\end{document}

%% file: preamble.tex
\usepackage{microtype}



%% file: sec/0_Abstract.tex
\begin{abstract}
Vision-language-action (VLA) models struggle to reuse recurring atomic actions across manipulation tasks. Our diagnostic study reveals that reliable task completion does not imply consistent execution of constituent atomic actions across task contexts. We present \textit{LexiconVLA}, a retrievable atomic-action lexicon for cross-task reuse. Global and detail codebooks capture shared interaction structure and fine-grained execution variation respectively, preserving both reusable patterns and execution details. Visual-Atomic Action Alignment couples trajectory reconstruction from visual state changes with visual outcome prediction from action codes, grounding the lexicon in motion and its effects. We learn these codebooks with trajectory reconstruction and visual alignment on our \textit{AtomAction} Dataset of 57{,}803 segments from 69 tasks. A planner and scene-aware adapter translate new goals into code-conditioned subtasks for a shared policy, without skill-specific experts or deployment-time parameter updates. Across five policy backbones on 26 RLBench tasks, \textit{LexiconVLA} largely maintains performance on 18 seen tasks while improving success on 8 tasks held out from policy training. With BridgeVLA, unseen-task success rises from 16.67\% to 34.17\% (+17.50 percentage points), and overall success reaches 71.08\%, the highest among methods with reported results. Real-robot experiments demonstrate stepwise execution and failure recovery. 
\end{abstract}

%% file: sec/1_Intro.tex
\section{Introduction}
\label{sec:intro}

VLA models enable instruction-guided manipulation by mapping visual observations and language instructions end-to-end to robot actions~\cite{brohan2023rt2visionlanguageactionmodelstransfer,kim2024openvlaopensourcevisionlanguageactionmodel,cheang2025gr3technicalreport,song2025reconvlareconstructivevisionlanguageactionmodel,black2026pi0visionlanguageactionflowmodel}.
Although scalable VLA frameworks yield powerful multi-task policies, reliable task completion does not necessarily imply that the underlying behaviors can be independently invoked and reused across task contexts. This limitation also matters for generalization, since novel tasks often involve interactions already encountered in prior manipulation experience. Essentially, diverse manipulation tasks share a set of fundamental behavioral primitives, termed atomic actions, including object approaching, grasping, and transferring, as shown in Figure \ref{fig:Teaser}. These basic interactive units recur across different tasks, despite variations in execution scenarios, target objects, and sequential arrangements. Accordingly, unseen manipulation tasks can be essentially regarded as new combinations of learned atomic actions. Based on this insight, the core challenge of cross-task generalization lies in the acquisition and structured organization of reusable behavioral primitives. This motivates a critical question: \textbf{How can prior manipulation experience be organized into reusable atomic-action representations that a unified policy can retrieve and compose across manipulation tasks?}

To investigate whether existing policies acquire reusable atomic actions, we conduct a diagnostic probe across five pretrained manipulation policies of diverse architectures across seven tasks, contrasting full-task success against the performance of each atomic constituent executed alone from its demonstration start state  (Appendix \ref{app:diagnostic}). High full-task success does not guarantee reliable atomic execution: across policies, task-averaged success drops from 65.7--95.7\% on complete tasks to 28.5--38.9\% on isolated atomic actions (Figure \ref{fig:Teaser}, bottom right), and even the common non-contact primitive approach varies widely across task contexts within a single policy (top right). These diagnostic results indicate that task-level training alone fails to yield atomic actions that can be reliably invoked independently—a capability needed for reuse across task contexts, including unseen tasks. This motivates atomic-level supervision and explicit action representations to retrieve shared interaction patterns while adapting to scene-specific conditions.


Recent atomic-skill methods~\cite{Zhang_2026_CVPR,sun2026atomvlascalableposttrainingrobotic,liu2026physicalwordsvocabularystructuring,shi2026atomicwamexecutiongroundedatomicmotionreasoning,wang2026skillmemoexpertguidedskillmemory} emphasize skill-specialized execution or subtask-level policy optimization using predicted outcomes as rewards. In contrast, we organize recurring interactions into representations that a shared policy can retrieve across tasks. To this end, \textit{LexiconVLA} learns an atomic-action codebook from interaction-phase trajectories and their observed visual effects. Quantization allows different trajectory instances to share a finite set of entries, providing a common action vocabulary rather than a separate representation for every instance. A scene-aware adapter retrieves these codes from the current observation and atomic subtask instruction, making trajectory-derived knowledge accessible without demonstrations at deployment. Retrieved codes condition a shared policy, and a planner sequences subtasks for new goals without skill-specific experts or deployment-time updates.

Learning a reusable action lexicon requires supervision that exposes segment-level structure shared across tasks, yet full demonstrations entangle multiple interaction phases and task-level instructions do not mark atomic-action boundaries. We therefore construct the \textit{AtomAction} Dataset, comprising 57,803 annotated atomic-action segments from 69 tasks, each paired with its trajectory, boundary observations, and language descriptions. To capture segment-level structure while retaining execution details, we use global and detail codebooks: a global code summarizes each segment, while detail codes provide complementary information for trajectory reconstruction. Reconstruction alone provides limited supervision about an action's effects on the environment. We thus introduce Visual-Atomic Action Alignment (V3A), coupling action reconstruction from visual changes with final-observation feature prediction, conditioned on the initial observation and the global action code. These objectives encourage action-relevant visual features and action codes predictive of observable effects, grounding the lexicon in both motion and its consequences.

Our contributions are threefold:
\begin{itemize}
\item We introduce \textit{AtomAction}, a dataset comprising 57,803 annotated atomic-action segments across 69 manipulation tasks. AtomAction pairs segmented interaction trajectories with boundary observations and textual descriptions. We further present a diagnostic analysis revealing a substantial gap between full-task completion and reliable standalone execution of constituent atomic actions.

\item We propose \textit{LexiconVLA}, a framework that distills recurring interaction phases into shared global and detail codebooks, grounded in their associated state transitions through Visual-Atomic Action Alignment (V3A). A planner schedules atomic subtasks, and a scene-aware adapter retrieves their codes to condition a pre-trained VLA. This closed loop composes atomic actions for manipulation tasks and enables recovery from execution failures.

\item Extensive experiments across five policy backbones show that LexiconVLA improves unseen-task success while largely preserving seen-task performance. With BridgeVLA, unseen-task success rises from 16.67\% to 34.17\% and overall success from 67.28\% to 71.08\%, the highest among evaluated methods. Real-robot rollouts further demonstrate sequential atomic-action execution and recovery from execution errors.
\end{itemize}

%% file: sec/2_Related_work.tex
\begin{figure*}[htbp!]
\vspace{-10pt}
\centering
\includegraphics[width=0.78\linewidth]{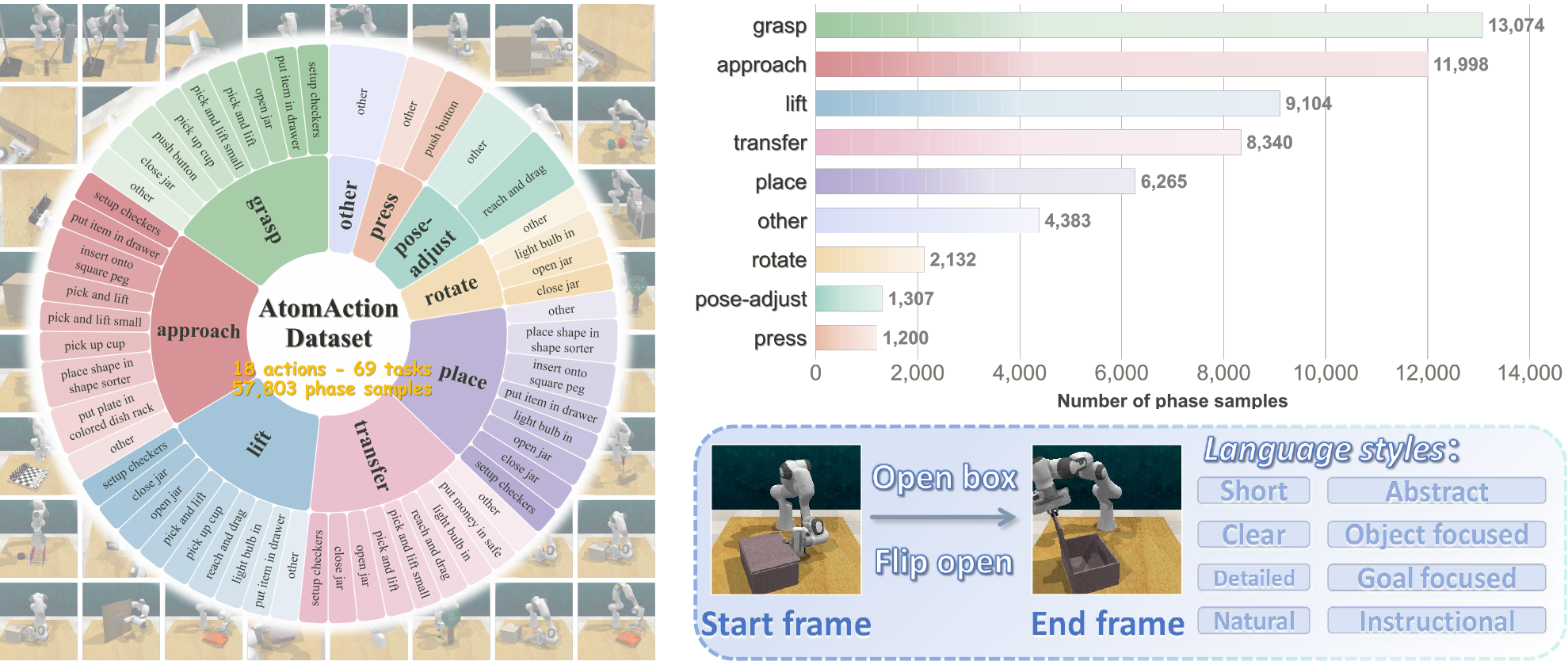}
\caption{\textbf{The AtomAction Dataset.} \textbf{Left:} Atomic-action categories (inner ring) and contributing tasks (outer ring). \textbf{Top right:} Segment counts per category; coloured block widths show individual task contributions, matching the outer ring. \textbf{Bottom right:} Example segment with start/end frames and language descriptions.}
\label{fig:AtomAction_Dataset_overview}
\vspace{-10pt}
\end{figure*}

\section{Related Work}

\subsection{Vision-Language-Action Models}
Vision-language-action (VLA) models build robot policies on pretrained vision-language models (VLMs) and large multi-robot datasets~\cite{octomodelteam2024octoopensourcegeneralistrobot,kim2024openvlaopensourcevisionlanguageactionmodel,intelligence2025pi05visionlanguageactionmodelopenworld}. Their action interfaces have evolved from autoregressively decoded discrete tokens~\cite{brohan2023rt2visionlanguageactionmodelstransfer,pertsch2025fastefficientactiontokenization,kim2025finetuningvisionlanguageactionmodelsoptimizing} to diffusion and flow-matching heads that generate continuous action chunks~\cite{chi2024diffusionpolicyvisuomotorpolicy,cheang2025gr3technicalreport,black2026pi0visionlanguageactionflowmodel,chen2026meanflowbasedonestepvisionlanguageaction}, while keyframe-based 3D policies attain strong multi-task manipulation performance~\cite{goyal2024rvt2learningprecisemanipulation,li2025bridgevla,li2026bridgevladataefficientgeneralizablememoryaugmented}. Across these designs, generalization derives largely from the pretrained backbone and training data. Although VLAs transfer to novel objects, scenes, and instruction phrasings, stronger general VLM capability does not reliably translate into better control~\cite{zhang2026vlm4vlarevisitingvisionlanguagemodelsvisionlanguageaction}, and perturbation studies indicate that high benchmark success can reflect memorized trajectories rather than task understanding~\cite{fei2025liberoplusindepthrobustnessanalysis,zhou2026liberoprorobustfairevaluation}. Recent work adds spatial representations~\cite{qu2025spatialvlaexploringspatialrepresentations,zhang20254dvlaspatiotemporalvisionlanguageactionpretraining,duisterhof2026pointzero3dpointtrack}, subtask reasoning~\cite{zawalski2025roboticcontrolembodiedchainofthought,zhao2025cotvlavisualchainofthoughtreasoning,puthumanaillam2026arcreasoningreciperobot}, and predictive world models~\cite{hu2025videopredictionpolicygeneralist,ye2026worldactionmodelszeroshot,zhou2026zerowamincontextworldactionmodeling,guo2026videopredictionpolicy2}, yet still learns the mapping from instruction to motion at the task level, leaving behaviors shared across tasks implicit in policy parameters and difficult to invoke independently. LexiconVLA instead makes these behaviors explicit as retrievable atomic-action codes that condition a shared policy.


\subsection{Learning Reusable Action Representations}
Existing action representations primarily differ in the unit they encode. Action tokenizers~\cite{lee2024behaviorgenerationlatentactions,zhou2025beastefficienttokenizationbsplines} discretize individual motions or fixed-length segments for reconstruction and generation fidelity, without explicit alignment to recurring atomic actions.
Latent action models infer actions from visual changes between unlabeled video frames~\cite{bruce2024geniegenerativeinteractiveenvironments,ye2025latentactionpretrainingvideos}, enabling pretraining across embodiments while also capturing task-irrelevant dynamics. Task-centric variants suppress these dynamics through language conditioning~\cite{bu2025univlalearningacttaskcentric,shi2026atomicwamexecutiongroundedatomicmotionreasoning}, but still represent short frame windows rather than complete interactions. Skill-level methods~\cite{Zhang_2026_CVPR,sun2026atomvlascalableposttrainingrobotic,li2026embodiedevoselfevolvingcodingagent,si2026fewertokensbetteraction} operate over longer horizons through specialized experts, agentic systems or subtask rewards during post-training. These skills are typically defined by goal-level labels that group mechanically distinct motions under terms such as \textit{open}, without shared representations for policy retrieval. LexiconVLA models interaction phases, supervised by trajectory reconstruction and state changes, to separate mechanisms under the same goal label and encode action effects.


%% file: sec/3_Dataset.tex
\section{AtomAction Dataset}

Learning a discrete vocabulary of atomic actions requires supervision at a granularity that existing manipulation corpora do not provide: most VLA datasets~\cite{9001253,NEURIPS2023_8c3c6668,zhou2026liberoprorobustfairevaluation,chen2026robodojounifiedsimandrealbenchmark} pair a complete demonstration with a single task instruction, so one trajectory spans several distinct interactions, while automatically derived atomic annotations remain coarse, a label such as \textit{open} subsuming translational pulling, outward hinge rotation, and lid flipping~\cite{Zhang_2026_CVPR,wei2026roboatomlargescaleatomicactionsupervision}. Segments sharing such a label stay heterogeneous in motion structure, and a codebook learned from them cannot separate mechanisms that a policy must execute differently. We therefore construct the \textbf{AtomAction Dataset}, in which every sample is a temporally localized segment realizing exactly one mechanism‑level atomic action. Built from RLBench~\cite{9001253} expert demonstrations, it comprises 57,803 segments from 69 tasks spanning 18 action types (Figure~\ref{fig:AtomAction_Dataset_overview}). Each sample retains the low‑dimensional state trajectory of the phase, five‑view RGB, depth, and mask observations at its boundaries, the atomic‑action label, and eight phase‑level descriptions in different language styles, illustrated by the \textit{flip-open} phase of \textit{open box} in the figure. The construction and annotation pipeline and atomic‑action standard are detailed in the Appendices~\ref{app:pipeline} and~\ref{app:action-standard}.

%% file: sec/4_Method.tex
\section{Method}


\begin{figure*}[t]
\vspace{-10pt}
\centering
\includegraphics[width=0.78\linewidth]{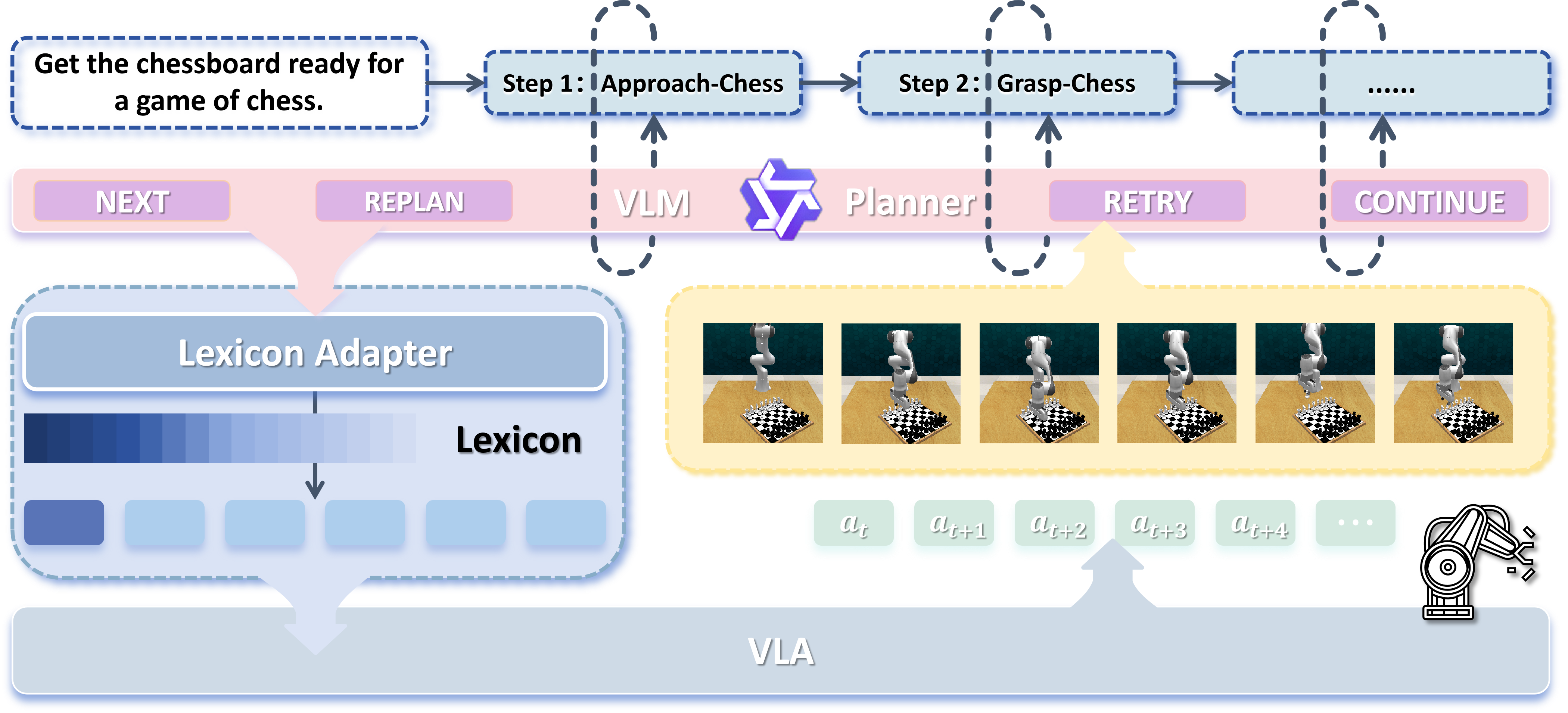}
\caption{\textbf{Overview of LexiconVLA.} A VLM-based planner decomposes the task instruction into atomic subtasks. Given the active subtask instruction and current observations, a scene-aware adapter retrieves entries from the frozen atomic-action lexicon---comprising global and detail codes---to condition a shared VLA policy. Execution feedback drives subtask progression and recovery through four skills: NEXT, CONTINUE, RETRY, and REPLAN. Subtask transitions and replanning refresh the retrieved codes based on updated observations.}
\label{fig:lexiconvla}
\vspace{-10pt}
\end{figure*}


\subsection{Overview}

\begin{figure*}[t]
\vspace{-10pt}
\centering
\includegraphics[width=0.78\linewidth]{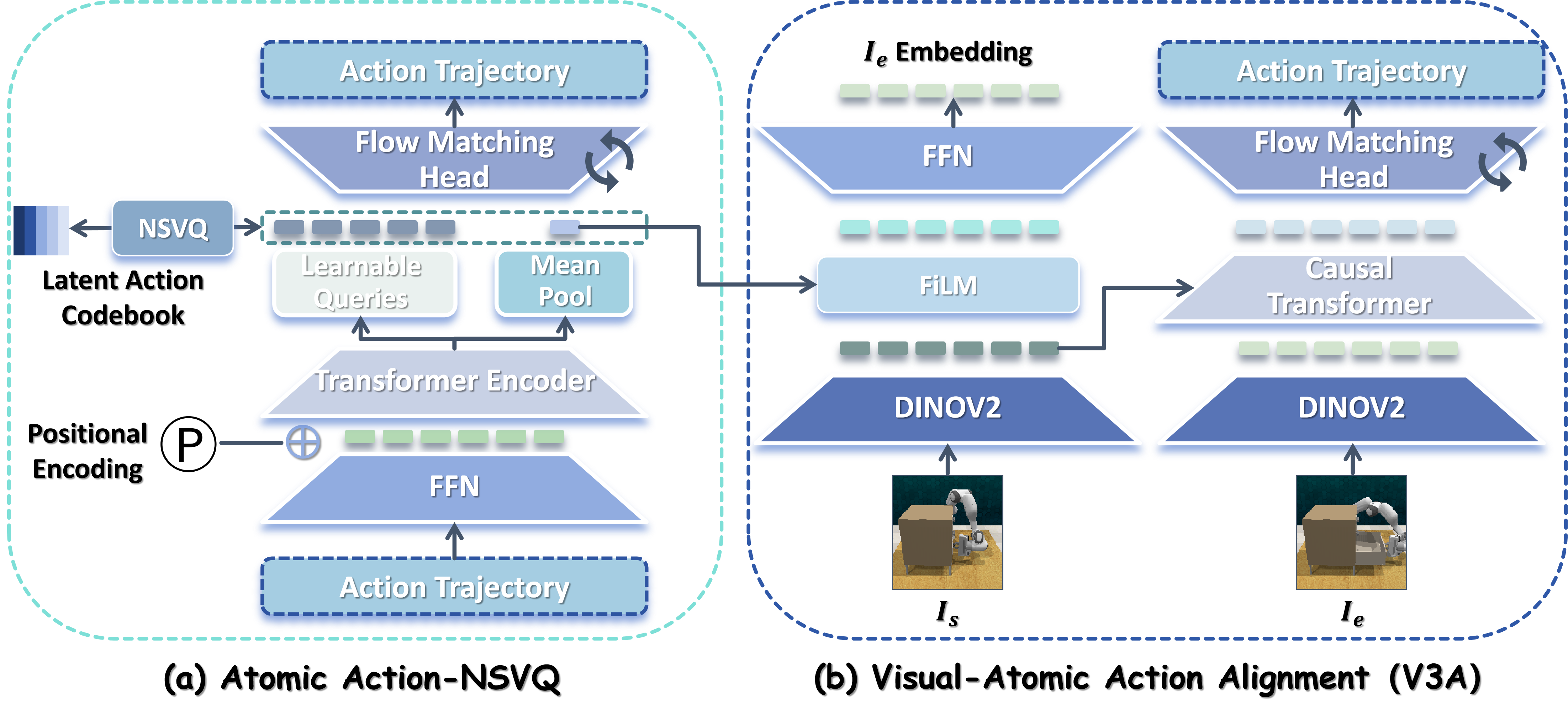}
\caption{\textbf{Overview of the Vector-Quantized Atomic Action Model (VQAA).} \textbf{(a)} Atomic Action-NSVQ encodes an atomic action segment into discrete codes from two learnable codebooks using Noise Substitution in Vector Quantization (NSVQ)~\cite{9696322}. A conditional flow matching decoder provides trajectory reconstruction supervision for learning the codebooks. \textbf{(b)} Visual-Atomic Action Alignment (V3A) incorporates visual context through two auxiliary objectives: reconstructing actions from initial-to-final visual changes and predicting final-frame features conditioned on the initial observation and the global action code.}
\label{fig:VQAA}
\vspace{-10pt}
\end{figure*}

LexiconVLA conditions a shared manipulation policy on a frozen atomic-action lexicon of global and detail codes (Figure~\ref{fig:lexiconvla}). We first describe execution (Sec.~\ref{sec:LexiconVLA}), then lexicon learning via the Vector-Quantized Atomic Action Model (VQAA; Sec.~\ref{sec:Vector-Quantized Atomic Action Model}, Figure~\ref{fig:VQAA}). VQAA integrates Atomic Action-NSVQ (Sec.~\ref{sec:Atomic Action-NSVQ}) for codebook learning from trajectory reconstruction and Visual-Atomic Action Alignment (V3A; Sec.~\ref{sec:Visual-Atomic Action Alignment}) to ground codes to observed state changes.

\subsection{LexiconVLA}
\label{sec:LexiconVLA}
Given the task instruction and current observations,
a VLM planner generates atomic subtask instructions.
A scene-aware adapter maps the active instruction $\ell_i$ and current images to codebook indices, retrieving $\mathbf z_g^{(i)}$ and $\mathbf Z_d^{(i)}$ as the corresponding lexicon entries. A lightweight injection module uses these codes to condition the shared VLA policy:
\begin{equation}
\mathbf b_t\sim\pi_\theta\!\left(
\cdot\mid O_t,\mathbf s_t,\ell_i,
\mathbf z_g^{(i)},\mathbf Z_d^{(i)}
\right),
\end{equation}
where $O_t$, $\mathbf s_t$, and $\mathbf b_t$ denote
visual observations, robot state, and the end-effector control command, respectively. The adapter learns this mapping using
code indices assigned to demonstration trajectories
by the frozen action encoder and codebooks
(Appendix~\ref{app:adapter-arc-train}).
Within a subtask, the retrieved codes stay fixed while the policy receives updated observations.
Based on visual feedback, the planner selects among four scheduling skills:
NEXT advances after subtask completion; CONTINUE keeps executing the active subtask;
RETRY repeats a failed step or returns to an earlier subtask;
and REPLAN updates the plan when scene changes invalidate it. See Appendices~\ref{app:lexiconvla-sch} and~\ref{app:planner-prompts} for monitoring rules and planner prompts.



\subsection{Vector-Quantized Atomic Action Model}
\label{sec:Vector-Quantized Atomic Action Model}


\subsubsection{Atomic Action-NSVQ}
\label{sec:Atomic Action-NSVQ}
\paragraph{Trajectory encoding.}
To capture differences in joint configuration, motion dynamics, and contact conditions that are not specified by a 6-DoF end-effector pose, we augment pose information with joint and gripper measurements, providing richer supervision for codebook learning. Let
$\mathbf A=[\mathbf a_1,\ldots,\mathbf a_T]^\top\in\mathbb R^{T\times39}$
denote an atomic action segment of length $T$ from AtomAction Dataset, where each timestep is represented as
\begin{equation}
\label{eq:atomic-action-input}
\begin{aligned}
\mathbf a_t=\bigl[&
\mathbf p_t,\; \mathbf r_t,\; \mathbf q_t,\; \dot{\mathbf q}_t,\\
&\mathbf f_t^{\mathrm{joint}},\;
\mathbf q_t^{\mathrm{grip}},\;
\mathbf f_t^{\mathrm{grip}},\; g_t
\bigr]^\top\in\mathbb R^{39}.
\end{aligned}
\end{equation}

Here, $\mathbf p_t\in\mathbb R^3$ and $\mathbf r_t\in\mathbb R^6$
encode end-effector position and rotation;
$\mathbf q_t,\dot{\mathbf q}_t,\mathbf f_t^{\mathrm{joint}}\in\mathbb R^7$
are joint positions, velocities, and torques;
$\mathbf q_t^{\mathrm{grip}}\in\mathbb R^2$,
$\mathbf f_t^{\mathrm{grip}}\in\mathbb R^6$, and $g_t\in\{0,1\}$
specify gripper joint positions, contact forces, and state.
After field-specific preprocessing
(Appendix~\ref{app:action-prep_norm}), we separately embed
the end-effector, joint, gripper-measurement, and gripper-state
groups and concatenate their embeddings. A channel encoder $E_{\mathrm{ch}}$ integrates these features through gating and feed-forward layers. A bidirectional Transformer~\cite{vaswani2017attention} $E_{\mathrm{temp}}^{\mathrm{RoPE}}$ equipped with RoPE~\cite{DBLP:journals/corr/abs-2104-09864} then models temporal dependencies:

\begin{equation}
\label{eq:trajectory-encoding}
\mathbf H
=E_{\mathrm{temp}}^{\mathrm{RoPE}}\!\left(
E_{\mathrm{ch}}\!\left(\Phi(\mathbf A)\right)
\right)
\in\mathbb R^{T\times d},
\end{equation}
where $\Phi$ denotes the grouped feature projections and $d$ is the hidden
dimension. RoPE~\cite{DBLP:journals/corr/abs-2104-09864} encodes relative temporal positions in self-attention to produce
contextualized features $\mathbf H=[\mathbf h_1,\ldots,\mathbf h_T]^\top$, with
$\mathbf h_t\in\mathbb R^d$. Padding positions are excluded throughout encoding
and aggregation.

\paragraph{Global and detail representations.}

Pooling $\mathbf H$ into a single vector may discard execution details needed for reconstruction.
We therefore complement a global feature obtained by masked average pooling with detail features extracted by learnable queries attending to $\mathbf H$:

\begin{equation}
\label{eq:global-detail-features}
\begin{aligned}
\bar{\mathbf h}_g
  &= \frac{\sum_{t=1}^{T}m_t\mathbf h_t}
           {\sum_{t=1}^{T}m_t}\in\mathbb R^d,\\
\bar{\mathbf H}_d
  &= \operatorname{CrossAttn}
     (\mathbf Q_d,\mathbf H,\mathbf H;\mathbf m)\in \mathbb R^{N_d\times d}.
\end{aligned}
\end{equation}


Here, $m_t\in\{0,1\}$ indicates a valid timestep, and $\mathbf Q_d\in\mathbb R^{N_d\times d}$ contains $N_d$ learnable queries.
Pooling summarizes the segment, while each detail query can emphasize different temporal information.
Separate linear projections map both feature types to a common code dimension $d_c$:

\begin{equation}
\label{eq:code-projections}
\begin{aligned}
\mathbf u_g&=P_g(\bar{\mathbf h}_g)\in\mathbb R^{d_c},\\
\mathbf U_d&=P_d(\bar{\mathbf H}_d)\in\mathbb R^{N_d\times d_c}.
\end{aligned}
\end{equation}

We quantize $\mathbf u_g$ using a global codebook
$\mathcal C_g$ and each row of $\mathbf U_d$ using a
shared detail codebook $\mathcal C_d$, yielding one global
code and $N_d$ detail codes per segment.

\paragraph{Noise-substituted vector quantization.}
To learn these discrete representations through reconstruction, we use Noise Substitution in Vector Quantization (NSVQ)~\cite{9696322}.
For a projected feature $\mathbf u\in\mathbb R^{d_c}$
and its corresponding codebook
$\mathcal C=\{\mathbf e_k\}_{k=1}^{K}$, the discrete
representation is the nearest codebook vector:
\begin{equation}
\label{eq:nearest-code}
k^\star=\underset{1\leq k\leq K}{\arg\min}
\|\mathbf u-\mathbf e_k\|_2^2,
\qquad
\mathbf z=\mathbf e_{k^\star}.
\end{equation}

To address the non-differentiability of nearest-neighbor selection, NSVQ replaces $\mathbf z$ with a noise-perturbed feature during training:
\begin{equation}
\label{eq:nsvq}
\tilde{\mathbf z}
=\mathbf u+
\frac{\|\mathbf u-\mathbf e_{k^\star}\|_2}
     {\|\boldsymbol\xi\|_2+\varepsilon}
\boldsymbol\xi,
\qquad
\boldsymbol\xi\sim\mathcal N(\mathbf0,\mathbf I_{d_c}),
\end{equation}
where $\varepsilon>0$ ensures numerical stability.
The noise scales with the quantization error, enabling reconstruction gradients to update both the encoder and the selected codebook vector without a straight-through estimator~(STE), compared with standard STE-based vector quantization~\cite{oord2018neuraldiscreterepresentationlearning,mentzer2023finitescalarquantizationvqvae}. Applying this operation to the two branches yields
$\tilde{\mathbf z}_g\in\mathbb R^{d_c}$ and
$\tilde{\mathbf Z}_d\in\mathbb R^{N_d\times d_c}$,
which jointly condition the trajectory decoder.

\begin{figure*}[!t]
\centering
\includegraphics[width=0.80\textwidth]{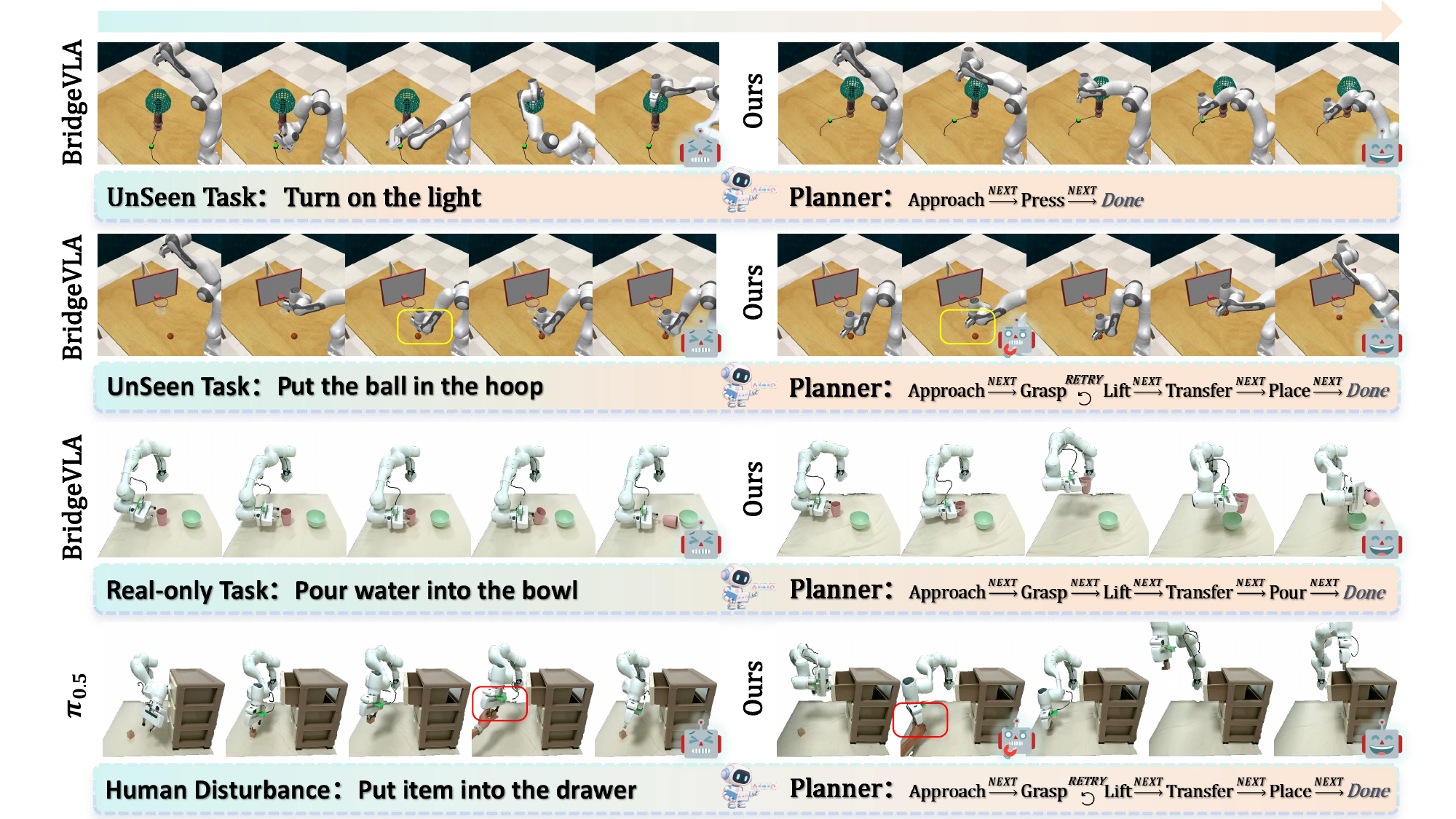}
\vspace{-8pt}
\caption{\textbf{Qualitative rollouts in simulation and the real world.} \textit{Ours}: LexiconVLA. Rows~1--2: policy-unseen RLBench tasks; rows~3--4: real-robot tasks. After a missed grasp (yellow box) or human disturbance (red box), our planner triggers RETRY, re-executing the failed subtask using the existing codes and updated observations, whereas the baselines fail to recover.}
\vspace{-10pt}
\label{fig:real_world_demos}
\end{figure*}

\paragraph{Conditional trajectory reconstruction.}
A conditional flow-matching decoder reconstructs
$\mathbf A_{\mathrm{ee}}\in\mathbb R^{T\times9}$,
comprising normalized positions and 6D rotations.
The global code and flow-time embedding modulate the decoder;
detail codes condition it through cross-attention.
For $s\sim\mathcal U(0,1)$ and Gaussian noise
$\boldsymbol\epsilon$, define
$\mathbf A_{\mathrm{ee}}^{(s)}
=(1-s)\boldsymbol\epsilon+s\mathbf A_{\mathrm{ee}}$
and target velocity
$\dot{\mathbf A}_{\mathrm{ee}}^{(s)}
=\mathbf A_{\mathrm{ee}}-\boldsymbol\epsilon$.
Following~\cite{lipman2023flowmatchinggenerativemodeling}, we minimize:


\begin{equation}
\begin{aligned}
\hat{\mathbf v}^{(s)}
&=v_\theta\!\left(
\mathbf A_{\mathrm{ee}}^{(s)},s
\mid\tilde{\mathbf z}_g,\tilde{\mathbf Z}_d
\right),\\
\mathcal L_{\mathrm{FM}}^{b}
&=\mathbb E_{\mathbf A,s,\boldsymbol\epsilon,\boldsymbol\xi}
\Bigl[
\bigl\|
\bigl[\hat{\mathbf v}^{(s)}-\dot{\mathbf A}_{\mathrm{ee}}^{(s)}\bigr]_b
\bigr\|_{\mathbf m}^{2}
\Bigr].
\end{aligned}
\end{equation}


where $b\in\{\mathrm{pos},\mathrm{rot}\}$,
$[\cdot]_b$ selects the corresponding position or rotation components, and $\|\mathbf X\|_{\mathbf m}^{2}
=\sum_t m_t\|\mathbf X_t\|_2^2/\sum_t m_t$
averages over valid timesteps.
A separate gripper head uses masked binary cross-entropy
$\mathcal L_{\mathrm{grip}}$, while
$\mathcal L_{\mathrm{sep}}
=\mathbb E[|\cos(\bar{\mathbf h}_g,
N_d^{-1}\sum_n\bar{\mathbf h}_{d,n})|]$
discourages redundancy between global and detail features,
with $\bar{\mathbf h}_{d,n}$ denoting the $n$-th detail feature.
The combined objective is
\begin{equation}
\mathcal L_{\mathrm{AA}}
=
\mathcal L_{\mathrm{FM}}^{\mathrm{pos}}
+\lambda_{\mathrm{rot}}\mathcal L_{\mathrm{FM}}^{\mathrm{rot}}
+\lambda_{\mathrm{grip}}\mathcal L_{\mathrm{grip}}
+\lambda_{\mathrm{sep}}\mathcal L_{\mathrm{sep}}.
\end{equation}
The decoder is used only during codebook training.


\subsubsection{Visual-Atomic Action Alignment}
\label{sec:Visual-Atomic Action Alignment}

Trajectory reconstruction captures motion but provides limited supervision on objects and action effects. Visual-Atomic Action Alignment (V3A) addresses this through action grounding from paired observations and final-frame feature prediction conditioned on the global code.

\paragraph{Action grounding from visual changes.}
Given the initial and final observations of an atomic action
segment from AtomAction Dataset, a shared DINOv2~\cite{oquab2024dinov2learningrobustvisual}
encoder extracts patch features
$\mathbf F_{\mathrm{start}},\mathbf F_{\mathrm{end}}
\in\mathbb R^{P\times d_v}$, where $P$ is the number of
patches and $d_v$ is the feature dimension.
We combine cross-attention with the projected feature
difference to encode visual changes:

\begin{equation}
\label{eq:visual-change-encoding}
\begin{aligned}
\mathbf F_\Delta ={}& E_\Delta\Bigl(
 \operatorname{CrossAttn}(\mathbf F_{\mathrm{end}}, \mathbf F_{\mathrm{start}},
 \mathbf F_{\mathrm{start}})\\
 &+P_\Delta(\mathbf F_{\mathrm{end}}
   -\mathbf F_{\mathrm{start}})\Bigr).
\end{aligned}
\end{equation}

Here, cross-attention uses final-frame features as queries
and initial-frame features as keys and values;
$P_\Delta$ projects the feature difference to the attention
output dimension, and $E_\Delta$ comprises two self-attention
blocks that produce the visual-change tokens $\mathbf F_\Delta$.

An independent decoder conditions on $\mathbf F_\Delta$ through cross-attention to reconstruct the action segment. It uses the same masked flow-matching losses for position and rotation and binary cross-entropy for gripper state as the trajectory reconstruction branch:
\begin{equation}
\mathcal L_{\mathrm{AG}}
=\mathcal L_{\mathrm{vis}}^{\mathrm{pos}}
+\lambda_{\mathrm{rot}}\mathcal L_{\mathrm{vis}}^{\mathrm{rot}}
+\lambda_{\mathrm{grip}}\mathcal L_{\mathrm{vis}}^{\mathrm{grip}}.
\end{equation}
The reconstruction weights are shared with $\mathcal L_{\mathrm{AA}}$. Gradients from this objective update the visual encoder through both observations via LoRA~\cite{hu2021loralowrankadaptationlarge}, encouraging features that capture action-relevant visual changes.

\paragraph{Global‑code‑conditioned visual prediction.}
To associate the global action representation with its visible effects, we predict final‑frame features from the initial observation and the global code $\tilde{\mathbf z}_g$, following the latent prediction principle of JEPA‑based world models~\cite{zhou2025dinowm,maes2026leworldmodelstableendtoendjointembedding,chen2026vljepajointembeddingpredictive}. The global code modulates initial‑frame features via FiLM before a shared MLP predicts each final‑frame patch:

\begin{equation}
\begin{aligned}
(\boldsymbol\gamma,\boldsymbol\beta)
&=P_{\mathrm{FiLM}}(\tilde{\mathbf z}_g),
\qquad \boldsymbol\gamma,\boldsymbol\beta\in\mathbb R^{d_v},\\
\hat{\mathbf f}_{\mathrm{end},p}
&=\operatorname{MLP}_{\mathrm{future}}\!\bigl((\mathbf 1+\boldsymbol\gamma)\odot\operatorname{LN}(\mathbf f_{\mathrm{start},p})
+\boldsymbol\beta\bigr)
\end{aligned}
\end{equation}

Here, $p=1,\ldots,P$, $P_{\mathrm{FiLM}}$ is a learned affine map, $\mathbf f_{\mathrm{start},p}$ is the initial feature at patch $p$, and $(\boldsymbol\gamma,\boldsymbol\beta)$ are shared across patches. We optimize a patch‑wise cosine loss:
\begin{align}
\mathcal L_{\mathrm{future}}
=\frac{1}{P}\sum_{p=1}^{P}
\left[
1-\cos\!\left(
\hat{\mathbf f}_{\mathrm{end},p},
\operatorname{sg}(\mathbf f_{\mathrm{end},p})
\right)
\right],
\end{align}

where $\operatorname{sg}$ denotes stop-gradient.
The initial-feature and global-code paths remain differentiable, propagating visual supervision to the action encoder and global codebook and encouraging global representations predictive of the observed state transition.

\paragraph{Training objective.}
VQAA combines trajectory reconstruction with the two visual objectives:
\begin{align}
\mathcal L_{\mathrm{VQAA}}
=\mathcal L_{\mathrm{AA}}
+\lambda_{\mathrm{AG}}\mathcal L_{\mathrm{AG}}
+\lambda_{\mathrm{future}}\mathcal L_{\mathrm{future}},
\end{align}
We gradually increase $\lambda_{\mathrm{future}}$ to strengthen visual supervision and set $\lambda_{\mathrm{AG}}=0$ once the visual encoder is frozen.

%% file: sec/5_Experiments.tex
\begin{table}[!t]
\small
\centering
\caption{\textbf{Multi-task performance on RLBench}~\cite{9001253}\textbf{.} Results are presented as mean±std success rate(\%) over three random seeds, with 25 evaluation episodes per task and seed. Shaded rows apply LexiconVLA to the base policy above. Best result per column in \textbf{bold}.  $^{*}$ denotes results we obtained by repeatedly fine-tuning and evaluating from the officially released pre-trained checkpoint.}
\vspace{-5pt}
\label{tab:success_rates}
\renewcommand{\arraystretch}{1}
\setlength{\tabcolsep}{2.5pt}
\scriptsize
\begin{tabular}{lc>{\hspace{-10pt}}lc>{\hspace{-10pt}}lc>{\hspace{-10pt}}l}
\toprule
\textbf{Method} & \makebox[0pt]{\textbf{Unseen (8 tasks)}} &  & \makebox[0pt]{\textbf{Seen (18 tasks)}} &  & \makebox[0pt]{\textbf{Overall (26 tasks)}} &  \\
\midrule
Act3D~\cite{gervet2023act3d3dfeaturefield} & $15.67_{\pm2.52}$ &  & $65.19_{\pm1.22}$ &  & $49.95_{\pm1.43}$ &  \\
3D Diffuser Actor~\cite{ke20243ddiffuseractorpolicy} & $28.67_{\pm0.29}$ &  & $78.30_{\pm1.92}$ &  & $63.03_{\pm1.34}$ &  \\
SAM2Act~\cite{fang2025sam2actintegratingvisualfoundation} & $15.50_{\pm1.50}$ &  & $84.30_{\pm1.80}$ &  & $63.13_{\pm1.02}$ &  \\
\midrule
PerAct~\cite{shridhar2022perceiveractormultitasktransformerrobotic} & $19.00_{\pm1.50}$ &  & $37.11_{\pm2.12}$ &  & $31.54_{\pm1.34}$ &  \\
\rowcolor{gray!10} \textbf{LexiconVLA (PerAct)} & $19.33_{\pm0.58}$ & \cellcolor{gray!10} & $39.48_{\pm2.11}$ & \cellcolor{gray!10} & $33.28_{\pm1.62}$ & \cellcolor{gray!10} \\
\midrule
RVT~\cite{goyal2023rvtroboticviewtransformer} & $11.83_{\pm2.31}$ &  & $59.33_{\pm2.04}$ &  & $44.72_{\pm2.00}$ &  \\
\rowcolor{gray!10} \textbf{LexiconVLA (RVT)} & $14.50_{\pm1.80}$ & \cellcolor{gray!10} & $60.81_{\pm1.14}$ & \cellcolor{gray!10} & $46.56_{\pm0.99}$ & \cellcolor{gray!10} \\
\midrule
RVT-2~\cite{goyal2024rvt2learningprecisemanipulation} & $18.00_{\pm0.50}$ &  & $75.63_{\pm1.00}$ &  & $57.90_{\pm0.64}$ &  \\
\rowcolor{gray!10} \textbf{LexiconVLA (RVT-2)} & $19.83_{\pm0.76}$ & \cellcolor{gray!10} & $76.30_{\pm0.34}$ & \cellcolor{gray!10} & $58.92_{\pm0.31}$ & \cellcolor{gray!10} \\
\midrule
BridgeVLA++$^{*}$~\cite{li2026bridgevladataefficientgeneralizablememoryaugmented} & $19.17_{\pm2.02}$ &  & $79.70_{\pm1.12}$ &  & $61.08_{\pm1.20}$ &  \\
\rowcolor{gray!10} \textbf{LexiconVLA (BridgeVLA++)} & $26.83_{\pm1.15}$ & \cellcolor{gray!10} & $77.04_{\pm1.68}$ & \cellcolor{gray!10} & $61.59_{\pm0.99}$ & \cellcolor{gray!10} \\
\midrule
BridgeVLA~\cite{li2025bridgevla} & $16.67_{\pm1.26}$ &  & $\mathbf{89.78}_{\pm1.02}$ &  & $67.28_{\pm0.94}$ &  \\
\rowcolor{gray!10} \textbf{LexiconVLA (BridgeVLA)} & $\mathbf{34.17}_{\pm2.02}$ & \cellcolor{gray!10} & $87.48_{\pm0.93}$ & \cellcolor{gray!10} & $\mathbf{71.08}_{\pm1.16}$ & \cellcolor{gray!10} \\
\bottomrule
\end{tabular}
\vspace{-10pt}
\end{table}

\section{Experiments}
\subsection{Simulation Experiments}
\paragraph{Experimental Setup.}
We evaluate on RLBench~\cite{9001253} with 18 seen and eight policy-unseen tasks. Although included in atomic-level pretraining, the eight policy-held-out tasks are excluded from downstream policy training, allowing us to evaluate the transfer of pretrained atomic knowledge to downstream task execution. Grouped by the atomic action they involve, these tasks span three categories: articulated-object manipulation (\textit{close drawer, open jar}), object transport and placement (\textit{pick up cup, phone on base, basketball in hoop, knife on board}), and switch operation (\textit{lamp on, press switch}). The codebooks and adapter are pretrained on AtomAction and subsequently frozen, while the policies are trained on demonstrations from the 18 seen tasks. At test time, the planner schedules subtasks and the adapter retrieves action codes from current observations and subtask instructions, without any parameter updates. We use Qwen3.8 as the default planner. We report full-task success over 25 episodes per task in each of three evaluation seeds. Appendix~\ref{app:sim-setup} details the training configurations, evaluation protocol and compared methods.

\paragraph{Results Analysis.}
Table~\ref{tab:success_rates} shows that strong seen-task performance of existing policies does not carry over to policy-unseen tasks: seven of the eight baselines reach only 11.83--19.17\% on unseen tasks, and BridgeVLA~\cite{li2025bridgevla}, the strongest on seen tasks (89.78\%), trails PerAct~\cite{shridhar2022perceiveractormultitasktransformerrobotic}, the weakest (37.11\%), on unseen tasks (16.67\% vs.\ 19.00\%). LexiconVLA mitigates this limitation across five backbones spanning voxel-based, multi-view, and VLM-based architectures. On every backbone, it improves unseen-task and overall success while largely preserving seen-task performance, even though it executes each task as a sequence of atomic subtasks whose success depends jointly on every subtask and transition. The unseen-task gains are largest on the two strongest backbones, BridgeVLA++ (+7.66 points)~\cite{li2026bridgevladataefficientgeneralizablememoryaugmented} and BridgeVLA (+17.50 points)~\cite{li2025bridgevla}. This trend is consistent with the division of labor in LexiconVLA: the retrieved codes specify which atomic action to perform, whereas the base policy executes it, so a more capable policy converts this guidance into task success more effectively.  With BridgeVLA~\cite{li2025bridgevla}, unseen-task success more than doubles from 16.67\% to 34.17\%, exceeding the strongest baseline, 3D Diffuser Actor (28.67\%)~\cite{ke20243ddiffuseractorpolicy}, by 5.50 points, with gains on six of the eight policy-held-out tasks (Appendix~\ref{app:per_task}). Overall success rises accordingly from 67.28\% to 71.08\%, the highest among all evaluated methods. Together, these results support combining pretrained atomic knowledge with subtask-conditioned execution as an effective approach to policy-unseen task transfer. Moreover, per-task results are reported in Tables~\ref{tab:sim-unseen} and~\ref{tab:sim-seen}.

\begin{table}[htbp]
  \centering
  \caption{Success rates (\%) in the ablation study.}
  \vspace{-5pt}
  \label{tab:module_ablation}
\scriptsize
    \setlength{\tabcolsep}{12pt}
  \renewcommand{\arraystretch}{1.08}
  \begin{tabular}{@{}lccc@{}}
    \toprule
    \textbf{Method} & \makecell{\textbf{Unseen}\\\textbf{(8)}}
      & \makecell{\textbf{Seen}\\\textbf{(18)}}
      & \makecell{\textbf{Overall}\\\textbf{(26)}} \\
    \midrule
    LexiconVLA (BridgeVLA) & \textbf{34.17} & 87.48 & \textbf{71.08} \\
    \quad w/o Planner            & 28.50 & \textbf{88.22} & 69.85 \\
    \quad w/o Codebooks          & 26.00 & 87.78 & 68.77 \\
    \quad w/o Global Codebook    & 26.50 & 85.33 & 67.23 \\
    \quad w/o Detail Codebook    & 29.50 & 87.11 & 69.38 \\
    \quad w/o V3A                & 26.50 & 84.89 & 66.92 \\
    \bottomrule
  \end{tabular}
  \vspace{-10pt}
\end{table}

\begin{table*}[!t]
\centering
\scriptsize
\caption{Real-world success rates (\%) on sim-to-real and generalization tasks.}
\vspace{-5pt}
\label{tab:sim2real_generalization}
\resizebox{\textwidth}{!}{
\begin{tabular}{c*{10}{c}}
\toprule
\multirow{5}{*}{\textbf{Methods}}
  & \multicolumn{5}{c}{\textbf{Standard}}
  & \multicolumn{5}{c}{\textbf{Generalization}} \\
\cmidrule(lr){2-6} \cmidrule(lr){7-11}
  & \multicolumn{2}{c}{\textbf{Sim-Short}}
  & \thead{\textbf{Sim-Long}}
  & \thead{\textbf{Real only-}\\\textbf{Short}}
  & \thead{\textbf{Real only-}\\\textbf{Long}}
  & \multicolumn{2}{c}{\textbf{Distractor}}
  & \thead{\textbf{Height}}
  & \thead{\textbf{Category}}
  & \thead{\textbf{Instruction}} \\
\cmidrule(lr){2-3} \cmidrule(lr){4-4} \cmidrule(lr){5-5} \cmidrule(lr){6-6}
\cmidrule(lr){7-8} \cmidrule(lr){9-9} \cmidrule(lr){10-10} \cmidrule(lr){11-11}
& \thead{Open\\drawer}
  & \thead{Put item\\in drawer}
  & \thead{Push\\buttons}
  & \thead{Pour water\\into the bowl}
  & \thead{Place vegetables\\on the plate}
  & \thead{Push\\buttons}
 & \thead{Place vegetables\\on the plate}
  & \thead{Pour water\\into the bowl}
   & \thead{Put item\\in drawer}
  & \thead{Open\\drawer} \\
\midrule
BridgeVLA~\cite{li2025bridgevla}         & 50.0 & 0.0 & 60.0 & 20.0 & 0.0 & 40.0 & 0.0 & 0.0 & 0.0 & 40.0 \\
$\pi_{0.5}$~\cite{intelligence2025pi05visionlanguageactionmodelopenworld}       & 60.0 & 0.0 & \textbf{80.0} & 0.0 & 0.0 & 40.0 & 0.0 & 0.0 & 0.0 & 20.0 \\
\rowcolor{gray!10}
LexiconVLA (Ours) & \textbf{100.0} & \textbf{40.0} & \textbf{80.0} & \textbf{40.0} & \textbf{50.0} & \textbf{60.0} & \textbf{20.0} & \textbf{80.0} & \textbf{40.0} & \textbf{80.0} \\
\bottomrule
\end{tabular}}
\vspace{-10pt}
\end{table*}

\subsection{Ablation and Analysis}

\begin{figure}[htbp]
\centering
\includegraphics[width=1\linewidth]{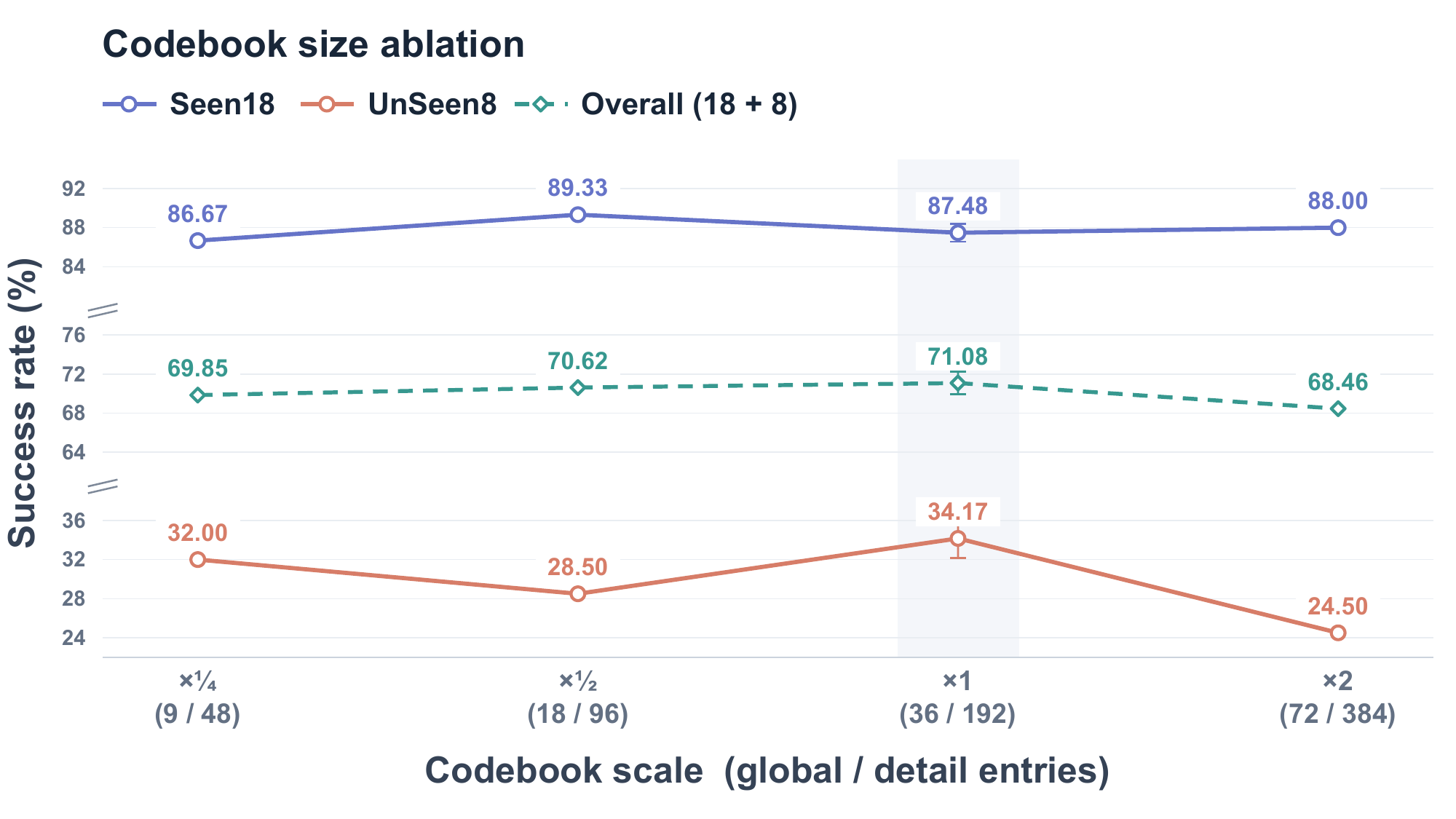}
\caption{\textbf{Codebook size ablation.} Success rates (\%) of LexiconVLA (BridgeVLA) with jointly scaled global/detail codebooks (entries in parentheses). The default size ($\times1$, shaded) reports mean$\pm$std over three runs; other sizes are evaluated once.}
\label{fig:codebook_size}
\vspace{-10pt}
\end{figure}

Table~\ref{tab:module_ablation} isolates the components of LexiconVLA with BridgeVLA; ablated variants are single runs, against a 2.02-point seed
deviation of the full model on policy-held-out tasks. Every ablation lowers
held-out success by at least 4.67 points while seen-task success stays within 3 points, so the components primarily serve transfer. Without the planner, codes are retrieved once from the full-task instruction and held-out success is 28.50\%, still 11.83 points above BridgeVLA; without the codebooks, the policy follows planner-generated subtasks alone and reaches 26.00\%. Decomposition thus explains part of the improvement but not all of it: code conditioning adds a further 8.17 points, and neither the lexicon nor the planner alone approaches the full model. The two codebooks are complementary rather than redundant: over the no-codebook control, global codes alone add 3.50 points and detail codes alone 0.50, whereas together they add 8.17, more than the sum. Removing V3A leaves held-out success at 26.50\%, on par with using no codes, showing that reconstruction alone does not yield transferable
codes and that supervision from observed state transitions is what makes the lexicon usable on new tasks. Finally, Figure~\ref{fig:codebook_size} scales
both codebooks jointly with full retraining: held-out success is highest at the default 36/192 entries among the tested sizes and drops most at $\times2$ (24.50\%), while all entries remain in use, so the differences reflect how interactions are partitioned rather than collapse. Analyses of code utilization, atomic-action information, and planner choice appear in Appendix~\ref{app:analysis}.

\subsection{Real-World Experiments}
\paragraph{Experimental Setup.}
We compare LexiconVLA (BridgeVLA) with its base policy BridgeVLA~\cite{li2025bridgevla} and the generalist VLA $\pi_{0.5}$~\cite{intelligence2025pi05visionlanguageactionmodelopenworld}, all fine-tuned on the same real-robot demonstrations per task. The \textit{Standard} setting evaluates five tasks under training conditions: three have RLBench counterparts (\textit{Sim}), and two have none (\textit{Real-only}). Following prior real-robot evaluations~\cite{li2025bridgevla,sun2026atomvlascalableposttrainingrobotic}, the \textit{Generalization} setting introduces distractors, elevated objects, unseen object categories, and paraphrased instructions, all absent from the demonstrations, without further training. Each task is evaluated over 10 trials per setting, and a trial succeeds only if the entire instruction is completed. Appendix~\ref{app:real-setup} details the hardware, tasks, perturbations, and evaluation protocol.

\paragraph{Results Analysis.}
Table~\ref{tab:sim2real_generalization} shows that LexiconVLA achieves the best or tied-best result under all ten conditions. Averaged over the five tasks, it reaches 62.0\% in the Standard setting versus 26.0\% for BridgeVLA~\cite{li2025bridgevla} and 28.0\% for $\pi_{0.5}$~\cite{intelligence2025pi05visionlanguageactionmodelopenworld}. The gain is most pronounced on multi-step tasks: both baselines fail every trial of put item in drawer and place vegetables on the plate, whereas LexiconVLA reaches 40.0\% and 50.0\%. Although the baselines open drawers in 50.0--60.0\% of open-drawer trials, they never complete drawer placement, mirroring the context-dependent atomic execution revealed by our diagnostic probe (Appendix~\ref{app:diagnostic}). Under perturbations absent from the demonstrations, LexiconVLA maintains 56.0\%, whereas BridgeVLA~\cite{li2025bridgevla} and $\pi_{0.5}$~\cite{intelligence2025pi05visionlanguageactionmodelopenworld} fall to 16.0\% and 12.0\%. Together with its lead on both real-only tasks, these results show that the atomic-action lexicon, learned solely from simulated segments, remains effective on a physical robot, both on tasks absent from simulation and under unseen conditions. In Figure~\ref{fig:real_world_demos}, the planner decomposes each instruction into atomic subtasks, such as approach, grasp, and transfer, which the policy executes with codes retrieved for each subtask, completing policy-unseen RLBench tasks and real-only pouring where BridgeVLA~\cite{li2025bridgevla} fails. Step-wise execution also localizes failures. After a failed grasp in put the ball in the hoop, or a human disturbance in put item into the drawer, the planner detects the failure from visual feedback, issues RETRY and resumes the failed action using updated observations and the existing codes. $\pi_{0.5}$~\cite{intelligence2025pi05visionlanguageactionmodelopenworld} has no explicit notion of which step has failed, and its rollout fails under the same disturbance. Appendix~\ref{app:real-analysis} details per-condition results and failure modes.

%% file: sec/6_Conclusion.tex
\section{Conclusion}
We presented LexiconVLA, which makes recurring manipulation interactions explicit and reusable across tasks. Motivated by our finding that task completion does not imply reliable atomic-action execution, we constructed the \textit{AtomAction} Dataset and learned global and detail codebooks grounded in motion and its visual effects through Visual-Atomic Action Alignment. A planner and a scene-aware adapter retrieve codes for each atomic subtask to condition a shared policy, without skill-specific experts or deployment-time parameter updates. Across five backbones on RLBench, LexiconVLA improves unseen-task success while largely preserving seen-task performance. On a real robot, the simulation-learned lexicon remains effective on tasks absent from simulation and under unseen conditions, and stepwise execution supports recovery from failures. Extending the lexicon with real-world and cross-embodiment data is a natural next step.





\clearpage
\subsection*{LLM Usage Statement}
We used generative AI to assist author-led work, not to conduct autonomous research, with a commitment to academic integrity and research ethics. Its uses included identifying relevant literature, drafting and polishing portions of the manuscript, checking clarity and consistency, and organizing existing data and experimental results into tables. For dataset construction, GPT-4o generated eight phase-level language descriptions for each annotated atomic-action segment, grounded in the action definition, task description, and local phase context (Appendices~\ref{app:pipeline} and~\ref{app:prompt}). These descriptions constitute the AI-generated language component of AtomAction. AI models also serve as components of the reported method and evaluation, including the VLM planner and the GPT-5.4 diagnostic judge, as described in Section~\ref{sec:LexiconVLA} and Appendix~\ref{app:diagnostic}. We also used Codex and Claude Code to assist us with code writing and code review.

We additionally used AI-assisted image editing to remove black curtain backgrounds from robot-manipulation photographs for clearer presentation of the robot and manipulated objects. We manually compared the edited images with the originals to verify that the depicted objects, robot actions, and task outcomes were preserved. 

Specifically, we used Claude Code and Codex for literature discovery and manuscript drafting and polishing, and Codex for AI-assisted background removal. For the VLM planner, we evaluated models from the Qwen family and providers including OpenAI. The compared planner models and their results are detailed in the ``Choice of planner'' analysis in Appendix~\ref{app:analysis} (Figure~\ref{fig:planner_ablation}).

All AI-assisted material underwent detailed human review before inclusion. We checked literature references and factual claims against original sources, checked tabulated values against the underlying data and experimental records, and reviewed generated language descriptions for consistency with their associated action segments and annotations. We also reviewed manuscript text for accuracy, attribution, and fidelity to our intended meaning, correcting or excluding unsupported, misleading, or fabricated content. AI output was not treated as evidence without verification. The authors retained responsibility for scientific judgments and conclusions and take full responsibility for the accuracy, originality, and integrity of the final manuscript, dataset, figures, and reported results.




%% file: sec/X_suppl.tex

\clearpage
\setcounter{page}{1}
\maketitlesupplementary

\appendix
\raggedbottom

\renewcommand{\thesection}{A.\arabic{section}}
\renewcommand{\thesubsection}{\thesection.\arabic{subsection}}

\titlecontents{section}
  [3em]{\bfseries}
  {\contentslabel{3em}}{}
  {\titlerule*[0.7pc]{.}\contentspage}
\titlecontents{subsection}
  [5.3em]{\normalfont}
  {\contentslabel{3.8em}}{}
  {\titlerule*[0.7pc]{.}\contentspage}

\startcontents[supp]
\section*{Contents}
\printcontents[supp]{}{1}{\setcounter{tocdepth}{1}}
\section{Diagnostic Probe of Atomic-Action Invocation}
\label{app:diagnostic}

\paragraph{Protocol.}
We probe five pretrained manipulation policies. We fix ten validation demonstrations per task for reach-and-drag, open-drawer, place-wine, sweep-to-dustpan, close-jar, push-buttons, and slide-block: 70 episodes, 290 segments, and 13 atomic labels. Selection requires complete keyframe coverage. Full-task evaluation uses RLBench success detection and 25 policy calls. For each isolated atom, we reset the demonstration, replay its ground-truth (GT) prefix, verify the handoff (position error $<2$\,cm and orientation error $<5^\circ$, matching gripper state). Invalid handoffs are excluded; segments are evaluated independently. GPT-5.4 judges front, wrist, and left-shoulder views with robot-state information, without policy identities. Strict scoring requires completed execution relative to the GT endpoint. Detailed results are provided in Table~\ref{tab:diagnostic}.


\begin{table*}[t]
\centering
\small
\setlength{\tabcolsep}{4pt}
\caption{Per-task results corresponding to Figure~\ref{fig:Teaser}. Each cell reports Full task SR / Atomic SR, in \%. Atomic SR pools valid segments within the task; full-task SR uses RLBench success detection.}
\label{tab:diagnostic}
\begin{tabular}{@{}lcccccc@{}}
\toprule
Task & PerAct & BridgeVLA++ & Act3D & RVT-2 & 3DDA & LexiconVLA \\
\midrule
Reach \& drag & 100.0 / 18.0 & 80.0 / 26.0 & 90.0 / 22.0 & 90.0 / 36.0 & 100.0 / 22.0 & 100.0 / 58.0 \\
Place wine & 50.0 / 29.7 & 90.0 / 33.3 & 80.0 / 26.3 & 80.0 / 45.0 & 80.0 / 25.6 & 70.0 / 69.2 \\
Sweep to dustpan & 70.0 / 0.0 & 100.0 / 40.0 & 100.0 / 17.5 & 100.0 / 17.5 & 80.0 / 25.0 & 100.0 / 35.0 \\
Slide block & 100.0 / 16.0 & 100.0 / 31.4 & 80.0 / 26.0 & 70.0 / 14.0 & 90.0 / 22.0 & 100.0 / 66.0 \\
Close jar & 40.0 / 14.0 & 100.0 / 25.0 & 90.0 / 24.0 & 20.0 / 20.0 & 60.0 / 0.0 & 100.0 / 67.5 \\
Push buttons & 20.0 / 40.0 & 100.0 / 56.7 & 100.0 / 50.0 & 100.0 / 40.0 & 100.0 / 66.7 & 100.0 / 83.3 \\
Open drawer & 80.0 / 81.5 & 100.0 / 60.0 & 90.0 / 65.5 & 90.0 / 76.7 & 100.0 / 64.3 & 100.0 / 83.3 \\
\bottomrule
\end{tabular}
\end{table*}


\section{Adapter Architecture and Training}
\label{app:adapter-arc-train}
\paragraph{Architecture.}
Frozen DINOv2, ViT‑B/14 and CLIP-ViT‑B/16 encoders tokenize the initial front/wrist images and subtask instruction. Pooling yields 64 visual tokens per camera, with learned spatial and camera embeddings. Together with 77 text positions, one global query, and $N_d=9$ detail queries, they form a 215‑token sequence. A four‑layer, 512‑dimensional Transformer with eight heads processes the sequence. Observation tokens attend only to observations; the global query also attends to itself; detail queries attend to all valid tokens. The final query features produce code logits:

\begin{equation}
\begin{aligned}
\mathbf o_g &= f_g(\mathbf h_g^{\mathrm{ada}})\in\mathbb R^{K_g},\\
\mathbf o_{d,n} &= f_d\!\left([\mathbf h_{d,n}^{\mathrm{ada}};
\mathbf h_g^{\mathrm{ada}}]\right)\in\mathbb R^{K_d},\\
&\quad n=1,\ldots,N_d.
\end{aligned}
\end{equation}

Here, $K_g=36$ and $K_d=192$ are the codebook sizes. Both heads are two‑layer GELU MLPs, with $f_d$ shared across detail slots. Argmax yields the predicted indices.

\paragraph{Training.}
Frozen Atomic Action‑NSVQ supplies nearest‑neighbor code indices from demonstration trajectories as fixed labels. The adapter learns to predict them from initial observations and instructions:


\begin{equation}
\begin{aligned}
\mathcal L_{\mathrm{adapter}}
={}&\mathbb E_{\mathcal D_{\mathrm{ada}}}\bigl[
\operatorname{CE}_{\eta}(\mathbf o_g,k_g^\star)\\
&+\frac{\lambda_d}{N_d}\sum_{n=1}^{N_d}
\operatorname{CE}_{\eta}(\mathbf o_{d,n},k_{d,n}^\star)
\bigr],\\
&\lambda_d=1,\quad\eta=0.05.
\end{aligned}
\end{equation}

where $\mathcal D_{\mathrm{ada}}$ denotes training data, starred indices are labels, and $\operatorname{CE}_{\eta}$ uses label smoothing $\eta$. The adapter is trained for 100 epochs with AdamW, batch size 64, and learning rate $3\times10^{-4}$, using three warmup epochs followed by cosine decay. The image/text encoders and codebooks remain frozen.

\paragraph{Deployment.}
The frozen adapter retrieves code vectors by predicted indices. CONTINUE and RETRY reuse the existing code vectors while the policy receives updated observations. RETRY does not invoke the adapter for fresh code retrieval.

\paragraph{Policy-side Code Injection.}
To incorporate the retrieved atomic-action representations into the base policy, we employ a lightweight conditioning module consisting of MLP projections and Feature-wise Linear Modulation (FiLM). The module maps the retrieved global and detail codes to conditioning parameters that modulate intermediate policy features through feature-wise scaling and shifting. We follow the same conditioning principle across the evaluated policy backbones, adapting the feature dimensions to their respective interfaces. This module serves as a simple interface between the learned action lexicon and the execution policy, rather than introducing additional skill-specific policy architectures.


\section{Additional Analyses}
\label{app:analysis}

\begin{figure*}[htbp]
\centering
\includegraphics[width=0.8\linewidth]{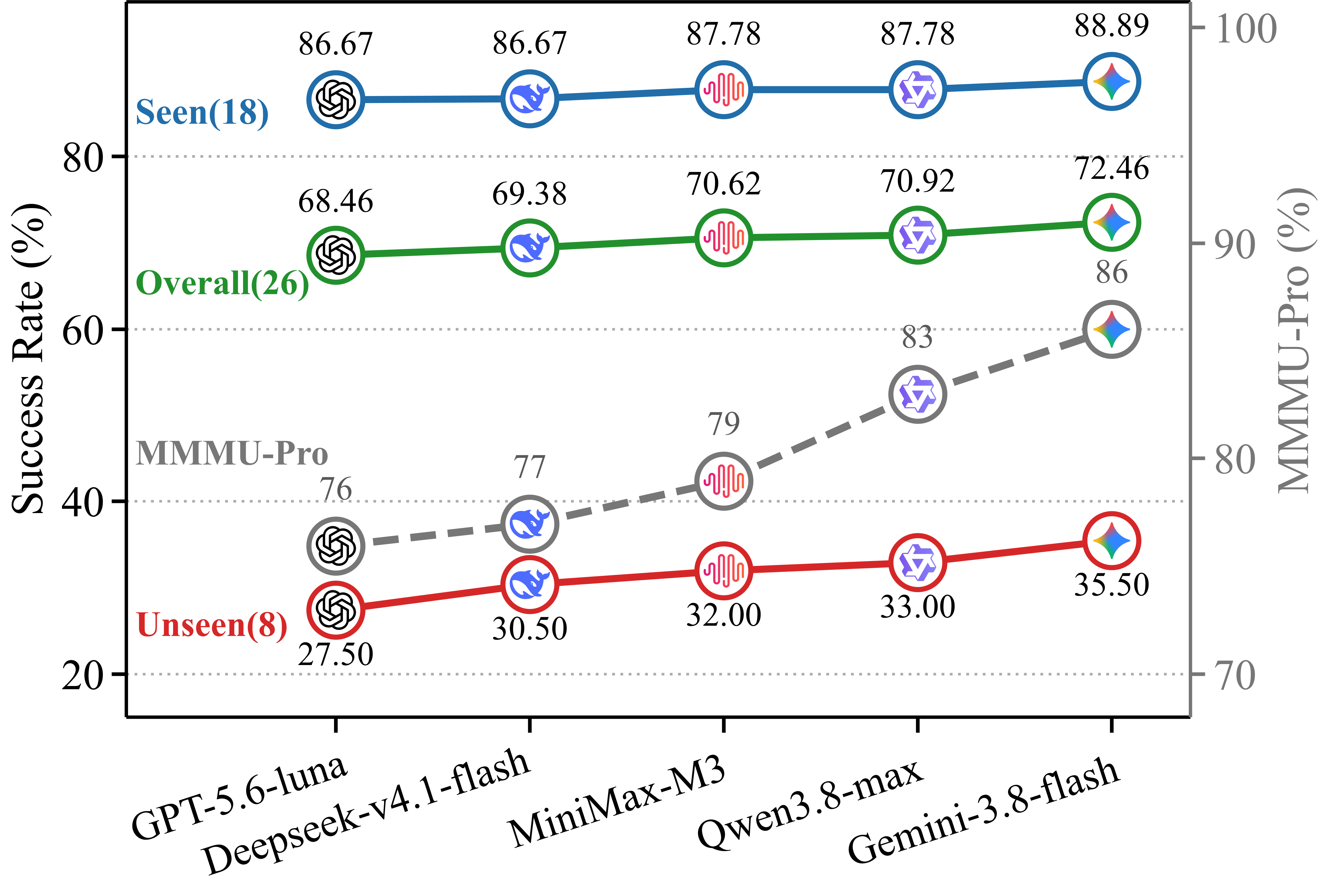}
\caption{\textbf{Effect of the planner model.} Success rates (\%) of LexiconVLA (BridgeVLA) with different planner models, ordered by MMMU-Pro score (dashed line, right axis). All other components are fixed; each planner is evaluated once.}
\label{fig:planner_ablation}
\end{figure*}

\paragraph{Choice of planner.}
We replace the model behind the planner with five alternatives while keeping the codebooks, adapter, policy (BridgeVLA), prompts, and evaluation episodes fixed (Figure~\ref{fig:planner_ablation}). Since the planner decomposes the task and monitors execution from visual observations, we relate success to each model's multimodal reasoning ability, measured by MMMU-Pro. Unseen-task success rises with MMMU-Pro from 27.50\% with gpt-5.6-luna to 35.50\% with gemini-3.8-flash, and overall success follows the same trend (68.46\% to 72.46\%), whereas seen-task success stays between 86.67\% and 88.89\%, consistent with seen tasks being largely handled by the base policy. Even the weakest planner improves unseen-task success over BridgeVLA by 10.83 points (16.67\% to 27.50\%), so the gains of LexiconVLA do not hinge on a particular planner, and stronger planners raise them further without retraining any component. 


\begin{figure}[t]
\centering
\includegraphics[width=0.85\linewidth]{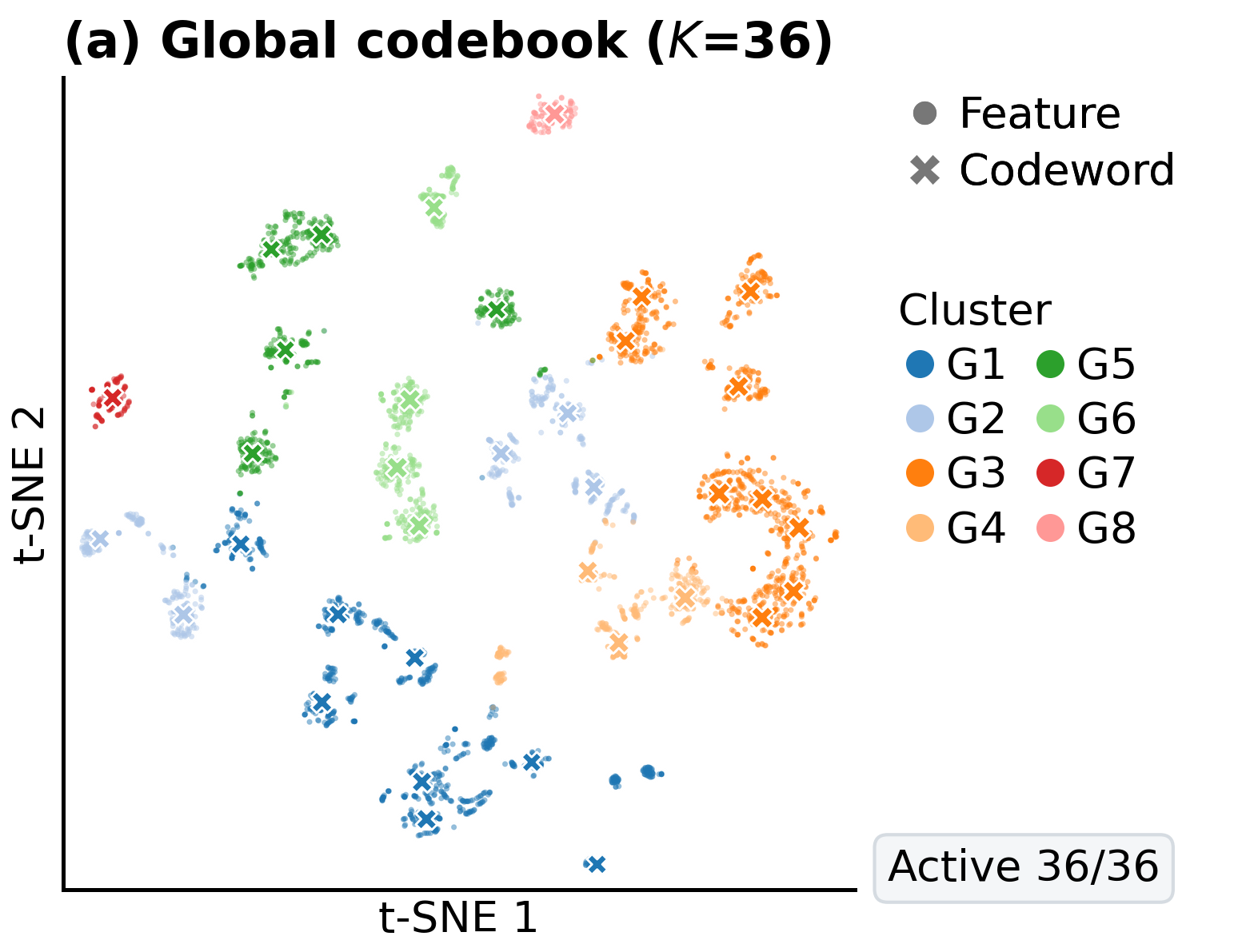}\\[4pt]
\includegraphics[width=0.85\linewidth]{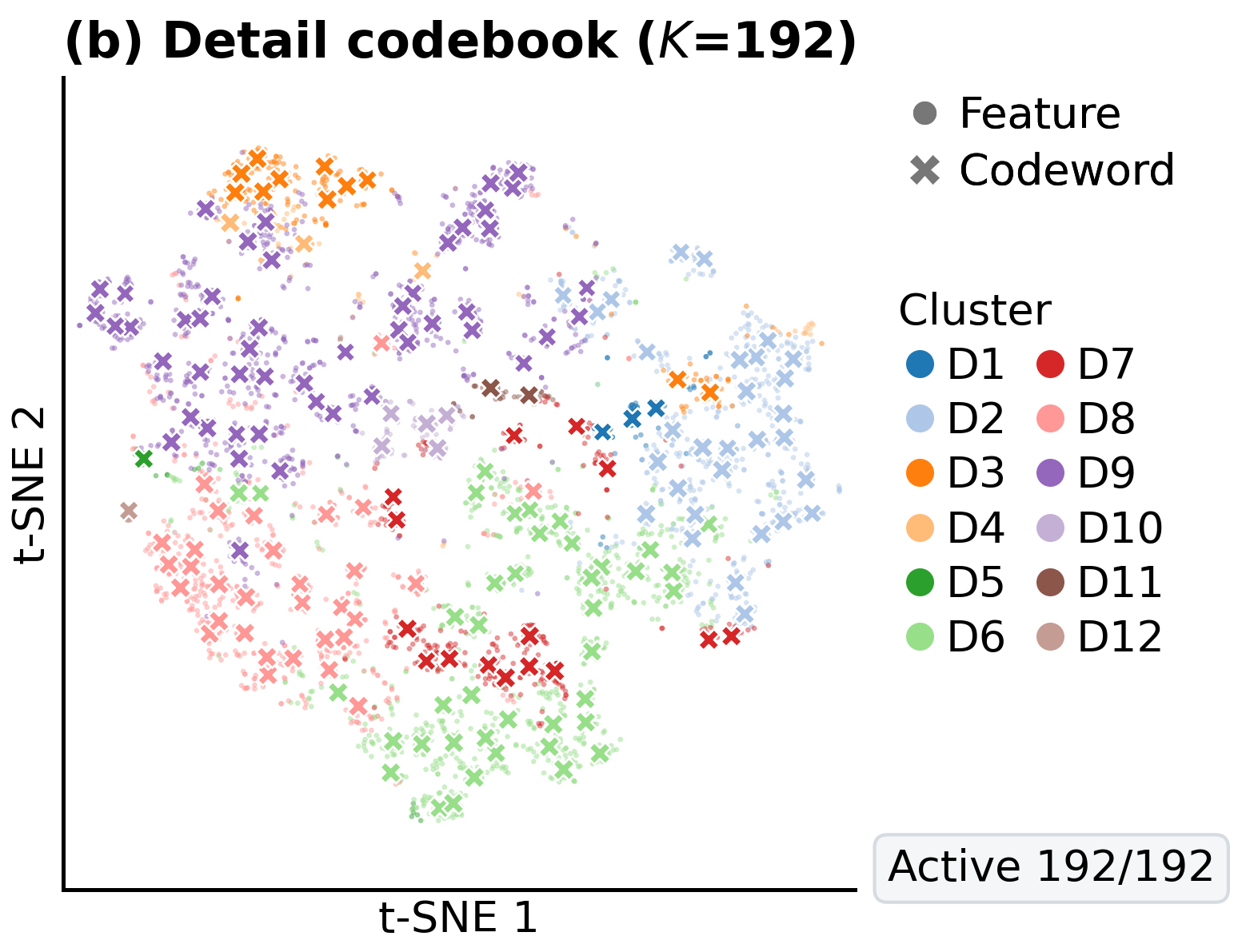}
\caption{\textbf{t-SNE of the learned (a) global and (b) detail codebooks.} Dots denote encoder features and crosses denote codewords; cross area scales with usage frequency.}
\label{fig:codebook_embedding}
\end{figure}

\paragraph{Learned codebook structure.}
We embed encoder features and all codewords of the default codebooks with t-SNE, coloring them by clusters of codeword vectors (Figure~\ref{fig:codebook_embedding}). All 36 global and 192 detail codewords are active, with normalized perplexities of 0.657 and 0.870, so neither codebook collapses onto a few entries. Global features form compact, well-separated groups around distinct codewords, indicating that the global codebook abstracts segments into a finite set of recurring patterns, while detail codewords densely cover a continuous region and capture the finer execution variation that complements the global code. To test whether the codes carry atomic-action information that generalizes across tasks, we fit a linear probe on the frozen codes with five-fold cross-validation grouped by task: on tasks held out from probe training, global and detail codes predict the annotated atomic action with 41.38\% and 31.08\% accuracy, respectively, compared with 23.58\% for the majority class. The codes thus retain atomic-action information that transfers across tasks, the property LexiconVLA relies on when retrieving codes for unseen tasks.


\section{LexiconVLA Scheduling}
\label{app:lexiconvla-sch}
\subsection{Closed-loop planning and execution}
\paragraph{Execution and monitoring.}
LexiconVLA coordinates task planning, lexicon-conditioned execution, and visual feedback through the closed-loop procedure in Algorithm~\ref{alg:lexiconvla}, drawing on feedback-driven orchestration frameworks such as OpenClaw and Hermes Agent. PLAN receives current images, the task instruction, and available decomposition and instruction priors, and produces atomic subtask instructions. A local state machine maintains the plan, active index, completion records, execution counters, and cached lexicon entries. The adapter retrieves codes for the active subtask, and the VLA executes end-effector commands using the latest observations, task/subtask instructions, and code context. After each keyframe action, LexiconVLA updates the execution state and checks whether monitoring is required. MONITOR receives the current images, plan, active subtask, elapsed steps, robot-state information, and trigger reason; it returns a scheduling decision, an optional target index, and a rationale. This permits selective VLM assessment while the policy continues to respond to observations.

\begin{algorithm}[htbp!]
\caption{LexiconVLA: online planning and execution}
\label{alg:lexiconvla}

\small
\noindent
\begin{tabularx}{\dimexpr\linewidth-20pt\relax}
    {@{}r@{\quad}>{\raggedright\arraybackslash}X@{}}
 & \textbf{Input:} Task instruction $\ell$, frozen codebooks,
   adapter, and VLA policy $\pi_\theta$. \\

1 & Observe $(O,\mathbf s)$;
    $P\leftarrow\mathrm{PLAN}(O,\ell)$;
    initialize active index $i\leftarrow1$. \\

2 & Initialize execution records and counters;
    initialize the retrieval key and code history. \\

3 & \textbf{while} the episode is active \textbf{do} \\

4 & \quad \textbf{if} the active subtask key $(i,\ell_i)$
    has changed and lexicon use is enabled \textbf{then} \\

5 & \qquad Predict codebook indices from $(O,\ell_i)$
    using the adapter. \\

6 & \qquad Retrieve and cache
    $(\mathbf z_g^{(i)},\mathbf Z_d^{(i)})$
    from the frozen lexicon; update the retrieval key
    and code history. \\

7 & \quad \textbf{end if} \\

8 & \quad Execute the policy with current observations,
    task/subtask instructions, and cached code context;
    disable injection if lexicon use is off. \\

9 & \quad Observe updated $(O,\mathbf s)$ and execution
    feedback; update counters. \\

10 & \quad Select the first eligible event or heartbeat
     trigger, if any. \\

11 & \quad \textbf{if} a trigger is present and monitoring
     is enabled within budget \textbf{then} \\

12 & \qquad \(\begin{aligned}[t]
     (a,j,r) &\leftarrow
     \mathrm{MONITOR}(O,\ell,P,i,\\
     &\quad \text{execution state},\text{trigger}).
     \end{aligned}\) \\

13 & \qquad Apply skill $a$ through the local state machine: \\

14 & \qquad \textbf{NEXT}: record completion and advance;
     extend the plan at its end. \\

15 & \qquad \textbf{CONTINUE}: retain the active subtask
     and cached codes. \\

16 & \qquad \textbf{RETRY}: retry step $j$ (default $i$);
     revoke rolled-back completions. \\

17 & \qquad \textbf{REPLAN}:
     $P\leftarrow\mathrm{PLAN}(O,\ell)$;
     reset $i$ and completion records. \\

18 & \qquad Apply recovery budgets and terminal
     plan-extension rules. \\

19 & \quad \textbf{end if} \\

20 & \textbf{end while} \\
\end{tabularx}

\vspace{2pt}
\noindent
\parbox{\dimexpr\linewidth-20pt\relax}{%
    \footnotesize
    $\ell_i$ is the instruction for the active atomic-action
    subtask in $P$; $j$ and $r$ are the optional target index
    and decision reason. The code context includes retrieved
    entries and causal global-code history. Trigger eligibility,
    code reuse, and budget-dependent overrides are specified below.
}
\end{algorithm}

\subsection{Monitoring and recovery rules}

\paragraph{Monitor triggers.}
Local checks run at each keyframe action step, rather than each simulation frame. In priority order, monitoring is triggered by:

\begin{table}[htbp]
\centering
\caption{Trigger conditions for invoking the MONITOR function}
\label{tab:monitor_triggers}
\begin{tabularx}{\linewidth}{@{}>{\raggedright\arraybackslash}p{0.29\linewidth}>{\raggedright\arraybackslash}X@{}}
\toprule
Trigger & Condition \\
\midrule
Gripper change & Binary gripper state changes, except for actions where closure serves as preparation or tool use. \\
Prolonged execution & The active subtask exceeds four keyframe steps. \\
Persistent inactivity & Inactivity persists for two steps, using joint‑speed and end‑effector displacement thresholds of $0.01$ rad/s and $0.005$ m; detected joint motion overrides the displacement test. \\
Heartbeat & Three keyframe steps have elapsed since the last monitoring call. \\
\bottomrule
\end{tabularx}
\end{table}


Only the first eligible trigger invokes MONITOR, with the heartbeat providing periodic assessment when no higher-priority event applies. At the final planned subtask, only prolonged execution and inactivity triggers remain active. A trigger requests assessment; it does not itself advance the plan.

\paragraph{Scheduling and recovery.}
The state machine applies the four core skills as follows:

\begin{table}[htbp]
\centering
\caption{State‑machine updates for each scheduling skill}
\label{tab:scheduling_skills}
\begin{tabularx}{\linewidth}{@{}>{\raggedright\arraybackslash}p{0.29\linewidth}>{\raggedright\arraybackslash}X@{}}
\toprule
Skill & State update \\
\midrule
\textbf{NEXT} & Mark the active subtask complete, advance by one step, and reset its execution and retry counters \\
\textbf{CONTINUE} & Retain the active subtask and cached codes \\
\textbf{RETRY} & Repeat the indicated failed step; when rolling back, revoke completion records from that step onward \\
\textbf{REPLAN} & Clear completion records and regenerate the plan from the current observation \\
\bottomrule
\end{tabularx}
\end{table}


NEXT requires visible completion. A reachable subtask with incomplete motion calls for CONTINUE; environmental changes that invalidate the plan call for REPLAN. RETRY can return to an earlier failed subtask and revoke subsequent completion records. These decisions use observed outcomes to guide progression and recovery.

\paragraph{Lexicon entry reuse.}
The current BridgeVLA integration retrieves codes when the active subtask key $(i,\ell_i)$ changes and lexicon use is enabled; otherwise, it reuses the cached entries. Disabling lexicon use clears the current codes, and episode reset clears the code history. A RETRY or REPLAN that retains the same index and instruction does not necessarily trigger fresh retrieval. Likewise, re-enabling lexicon use without changing this key does not independently trigger retrieval. These are implementation limits of the current cache, rather than guarantees of scene-dependent retrieval after every recovery decision.

\paragraph{Plan exhaustion and limits.}
The implementation additionally supports EXTEND: when the final planned step is exhausted, a final‑step NEXT is converted into a request for remaining subtasks. EXTEND preserves progress and appends the new steps. Extensions are limited to two, with at most ten planned segments; an empty or blocked extension disables further monitoring. Each episode permits at most 25 VLM calls and four replans. Further replan requests become CONTINUE. After more than four retries, the controller switches the decomposition prior where available and advances beyond the repeatedly failed step. These limits bound planner usage and recovery attempts.


\section{Action Preprocessing and Normalization}
\label{app:action-prep_norm}

The encoder input contains 39 scalar components: 3 for end-effector position,
6 for rotation, 21 for joint positions, velocities, and torques, 8 for gripper
joint positions and contact forces, and 1 for the binary gripper state. These
components describe complementary aspects of the recorded robot state. In
particular, the six rotation components encode a three-degree-of-freedom
orientation; the input dimension does not denote the robot's number of degrees
of freedom.

\paragraph{Field-specific normalization.}
\begin{table}[htpb]
\centering
\caption{Preprocessing of the 39-dimensional encoder input.}
\label{tab:action-preprocessing}
\small
\renewcommand{\arraystretch}{1.15}
\begin{tabular}{@{}p{0.38\linewidth}cp{0.43\linewidth}@{}}
\hline
\textbf{Input field} & \textbf{Dim.} & \textbf{Preprocessing} \\
\hline
End-effector position $\mathbf p_t$ & 3 & Quantile normalization \\
End-effector rotation $\mathbf r_t$ & 6 & Quaternion-to-6D conversion; no statistical normalization \\
Joint positions $\mathbf q_t$ & 7 & Z-score normalization \\
Joint velocities $\dot{\mathbf q}_t$ & 7 & Quantile normalization \\
Joint torques $\mathbf f_t^{\mathrm{joint}}$ & 7 & Quantile normalization \\
Gripper joint positions $\mathbf q_t^{\mathrm{grip}}$ & 2 & Z-score normalization \\
Gripper contact forces $\mathbf f_t^{\mathrm{grip}}$ & 6 & Quantile normalization \\
Gripper state $g_t$ & 1 & Binary embedding; no statistical normalization \\
\hline
\end{tabular}
\end{table}

We normalize each scalar dimension independently using dataset-level statistics
shared across all action categories. The current preprocessing computes these
statistics from all indexable trajectories under the dataset root, before
sequence padding. The transformations are summarized in
Table~\ref{tab:action-preprocessing}.

\begin{enumerate}
  \item For a scalar measurement $s$, quantile normalization uses its first and
ninety-ninth percentiles, denoted by $Q_{0.01}$ and $Q_{0.99}$:
\begin{equation}
\label{eq:quantile-normalization}
\begin{aligned}
\tilde s &= 2\frac{s-Q_{0.01}}
{\max(Q_{0.99}-Q_{0.01},\epsilon_{\mathrm{norm}})}-1,\\
\epsilon_{\mathrm{norm}} &= 10^{-6}.
\end{aligned}
\end{equation}
This scaling limits the influence of extreme observations on the normalization
range. The percentile endpoints map to $-1$ and $1$ whenever their separation is
at least $\epsilon_{\mathrm{norm}}$. We do not clip the normalized values, so
observations outside the percentile interval retain their relative magnitudes
and may fall outside $[-1,1]$.

  \item Joint and gripper joint positions are standardized using their per-dimension
mean $\mu$ and population standard deviation $\sigma$:

\begin{equation}
\label{eq:zscore-normalization}
\tilde s=\frac{s-\mu}{\max(\sigma,\epsilon_{\mathrm{norm}})}.
\end{equation}

\end{enumerate}

Both transformations impose a lower bound on the denominator for numerical stability. The normalization statistics are fixed during model training, and the same position transformation is used for encoder inputs and trajectory reconstruction targets.

\paragraph{Rotation and gripper representations.}
The recorded end-effector quaternion is converted to a rotation matrix
$\mathbf R_t\in\mathrm{SO}(3)$. We concatenate its first two columns to obtain
\begin{equation}
\label{eq:rotation-6d}
\mathbf r_t=
\left[(\mathbf R_t)_{:,1}^{\top},\;
(\mathbf R_t)_{:,2}^{\top}\right]^{\top}
\in\mathbb R^6.
\end{equation}
These components lie in $[-1,1]$ and receive no further statistical
normalization. The binary gripper state $g_t$ is mapped to a learnable
64-dimensional embedding before its feature projection. Padding is introduced
after preprocessing, and the resulting validity mask excludes padded timesteps
from encoding, aggregation, and reconstruction losses.
\section{Dataset Construction and Annotation Pipeline}
\label{app:pipeline}



\begin{figure*}[t]
    \centering
    \includegraphics[width=\linewidth]{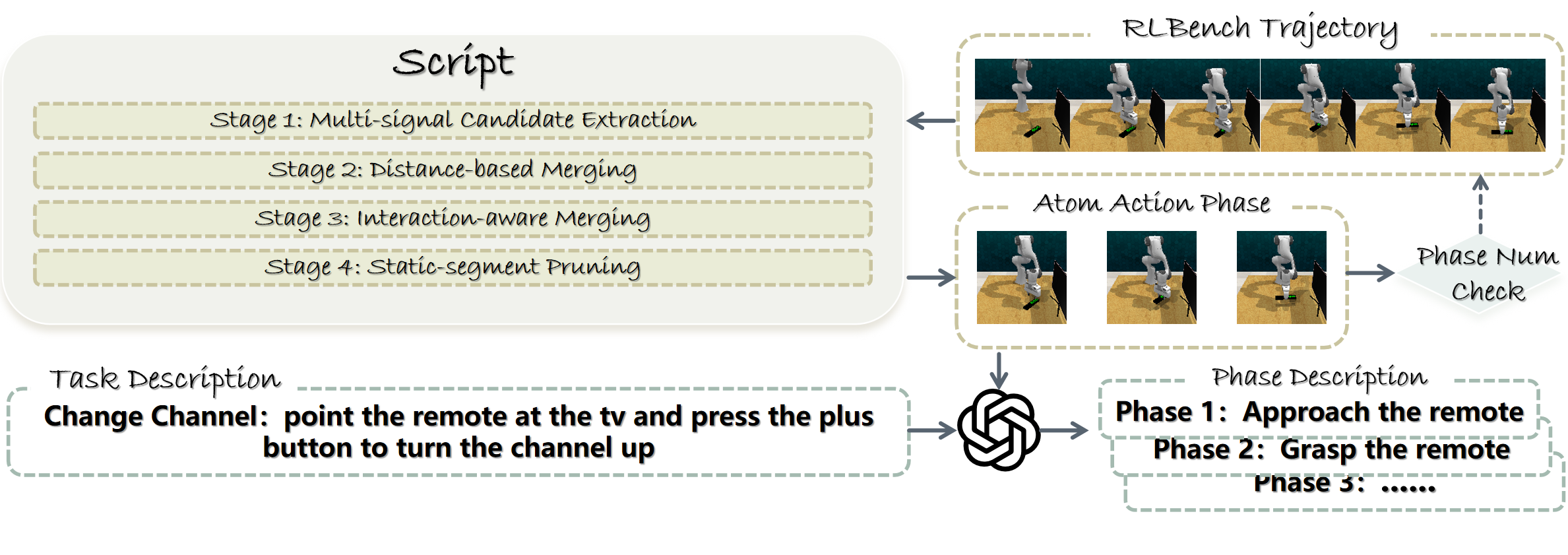}
    \caption{
    AtomAction Dataset construction pipeline. Complete RLBench expert
    trajectories are segmented by a rule-based procedure, checked against
    expert-provided phase-count priors, and annotated with atomic-action
    labels and phase-level language descriptions.
    }
    \label{fig:appendix_pipeline}
\end{figure*}

Figure~\ref{fig:appendix_pipeline} presents the construction pipeline of
the AtomAction Dataset. Starting from complete RLBench expert
trajectories, the pipeline first applies a rule-based script to divide
each continuous demonstration into temporally localized manipulation
phases. The script performs four successive operations: multi-signal
candidate extraction, distance-based merging, interaction-aware merging,
and static-segment pruning.

\paragraph{Stage 1: Multi-signal candidate extraction.}
The script identifies candidate transition points from complementary
trajectory signals, including gripper-state changes, velocity stop/start
events, and contact-related changes. The union of these events forms an
inclusive set of candidate phase boundaries.

\paragraph{Stage 2: Distance-based merging.}
The initial candidate set may contain several boundaries around the same
physical transition. To reduce such over-segmentation, adjacent candidates
separated by fewer than five frames are merged, with the later boundary
retained. This operation suppresses short fragments caused by multiple
signals responding to the same transition.

\paragraph{Stage 3: Interaction-aware merging.}
The resulting intervals are further refined using contact, gripper,
motion, and torque-related information. Consecutive intervals belonging
to the same object interaction are merged into a coherent phase.
Direction-aware processing is also applied to non-interaction intervals:
approach and intermediate transit motion are preserved when relevant,
whereas task-completion reset motion can be removed.

\paragraph{Stage 4: Static-segment pruning.}
Intervals containing a stationary run of at least eight frames are
removed unless they still exhibit meaningful accumulated positional or
rotational changes. This step suppresses low-information pauses while
retaining slow but purposeful manipulation. The final retained interval
is extended when necessary to cover the end of the episode.

\paragraph{Phase-count checking and refinement.}
For each task variation, experts first inspect representative
demonstration videos and use task-specific knowledge to determine the
expected number of phases in its episodes. This expert-provided phase-count
prior is incorporated into the automated pipeline as an acceptance
criterion. After each segmentation attempt, the script compares the
detected phase count with the corresponding prior; a mismatch automatically
triggers another segmentation cycle, up to the configured retry limit.
This iterative process reduces over- and under-segmentation caused by
transient gripper, motion, contact, or collection noise. Most episodes can
therefore be processed automatically, while only the remaining ambiguous
cases require lightweight manual boundary refinement, substantially
improving segmentation reliability with limited annotation effort.
After the phase boundaries are finalized, experts assign an atomic-action
type to each phase according to the definitions in
Appendix~\ref{app:action-standard}.

\paragraph{Language annotation and dataset assembly.}
Each annotated phase segment and its corresponding atomic-action
definition are combined with the task description and local phase context
to form the language-generation prompt. GPT-4o then generates eight
phase-specific English descriptions for every atomic-action phase.
The complete prompt structure and an example output are provided in Appendix~\ref{app:prompt}.

Finally, the annotated phases are reorganized into an action-first
hierarchy indexed by atomic action, task, variation, and phase. Each
AtomAction sample contains the complete low-dimensional robot-state
trajectory of the phase; five-view RGB, depth, and mask observations at
the phase boundaries; task and variation information; the atomic-action
label and phase metadata; and eight phase-level English descriptions.
Together, these components constitute the final AtomAction Dataset.


\Needspace{0.42\textheight}\section{Comparison with AtomicVLA Data}
\label{sec:atomicvla-comparison}


\begin{figure*}[t]
    \centering
    \includegraphics[width=\linewidth]
    {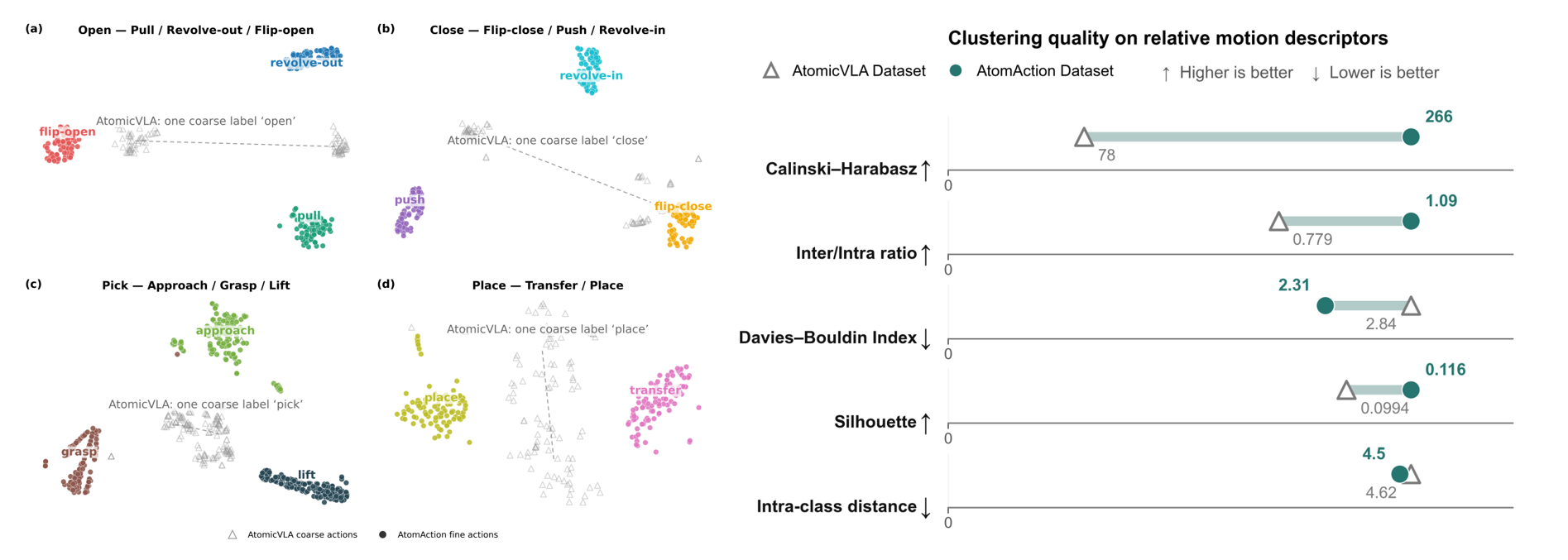}
    \caption{
    Motion-structure comparison between AtomicVLA and AtomAction.
    Left: four action families. Right: clustering metrics.
    Gray triangles denote AtomicVLA coarse labels; colored circles
    denote AtomAction fine-grained labels.
    }
    \label{fig:comparison}
\end{figure*}

AtomicVLA constructs atomic-action annotations by analyzing relative
end-effector translation, rotation, and gripper changes within short
motion windows. Principal-axis analysis yields
initial temporal segments and coarse labels, which are then refined from
video clips using InternVideo2.5. In its LIBERO setting, trajectories are
organized into five action abstractions: Pick, Place, Open, Close, and
Turn. Each segment retains its visual observations, robot actions, task
instruction, temporal interval, and coarse label.

However, this representation remains coarse-grained. A single label often
mixes different motion mechanisms or execution stages. For example, Open
combines translational pulling, outward hinge rotation, and lid opening,
whereas Pick combines approaching, grasping, and lifting. Segments sharing
an AtomicVLA label therefore remain heterogeneous in motion structure, limiting annotation precision at the mechanism level. AtomAction addresses
this with 18 phase-level action types, including Pull, Revolve-out,
Flip-open, Approach, Grasp, Lift, Transfer, and Place.

We further conduct a quantitative comparison. From the Open, Close, Pick,
and Place families in both datasets, we extract independent trajectory
segments, convert them into 32-step, seven-channel relative end-effector
sequences, and construct 52-dimensional motion descriptors. The left panels illustrate how AtomAction distinguishes motion mechanisms and execution stages within coarse action families: Open separates into Pull, Revolve-out, and Flip-open, while Pick
separates into Approach, Grasp, and Lift. These finer labels correspond
to distinct motion groups in the visualization. Consistent with this
pattern, the quantitative results on the right show higher
Calinski--Harabasz (266 vs.\ 78), inter-/intra-class ratio
(1.09 vs.\ 0.779), and silhouette (0.116 vs.\ 0.0994) scores for
AtomAction, together with a lower Davies--Bouldin index
(2.31 vs.\ 2.84) and a modest reduction in intra-class distance
(4.50 vs.\ 4.62). These results indicate greater between-class
separability and slightly improved within-class compactness under
the selected subsets and shared motion representation.

Taken together, these findings suggest that, compared with AtomicVLA's
coarse action annotations, AtomAction provides finer distinctions
between manipulation mechanisms and execution stages, with labels
that better reflect the structure of the underlying motion.
This finer granularity supports more precise grouping and retrieval
of atomic actions and offers a basis for aligning related actions
across tasks. AtomAction thus provides a structured source of
data for future research on atomic action representation
learning, cross-task skill transfer, and compositional robot learning.


\section{Simulation Setup and Evaluation Protocol}
\label{app:sim-setup}

\paragraph{Task Split and Data Coverage.}
The seen set consists of the standard 18 RLBench tasks used
for policy fine-tuning.
The policy-unseen set contains eight additional tasks excluded
from policy training. This setting evaluates whether pretrained atomic knowledge
can support downstream execution of policy-unseen tasks.
Table~\ref{tab:sim-task-split} lists the task groups; policy-unseen tasks are grouped by the atomic action they involve.


\begin{table}[t]
\centering
\small
\caption{\textbf{Simulation task groups and training coverage.}}
\label{tab:sim-task-split}
\renewcommand{\arraystretch}{1.15}
\begin{tabularx}{\linewidth}{@{}
  >{\raggedright\arraybackslash}X
  >{\raggedright\arraybackslash}X
  >{\raggedright\arraybackslash}X@{}}
\toprule
\multicolumn{3}{@{}l}{\textbf{Seen tasks (18)}} \\
\midrule
Close jar & Light bulb in & Open drawer \\
Place cups & Place shape in shape sorter & Push buttons \\
Put groceries in cupboard & Reach and drag & Slide block to color target \\
Stack blocks & Place wine at rack location & Sweep to dustpan of size \\
Insert onto square peg & Meat off grill & Put item in drawer \\
Put money in safe & Stack cups & Turn tap \\
\midrule
\multicolumn{3}{@{}l}{\textbf{Policy-unseen tasks (8)}} \\
\midrule
\textit{Articulated-object manip.\ (2)} &
\textit{Object transport \& placement (4)} &
\textit{Switch operation (2)} \\
\cmidrule(r){1-1}\cmidrule(lr){2-2}\cmidrule(l){3-3}
Close drawer & Pick up cup & Lamp on \\
Open jar & Phone on base & Press switch \\
           & Basketball in hoop & \\
           & Put knife on chopping board & \\
\bottomrule
\end{tabularx}
\end{table}

\paragraph{Atomic Pretraining and Policy Adaptation.}
We pretrain the codebooks and adapter on AtomAction and
freeze both during downstream policy adaptation.
The codebooks contain 36 global entries and 192 detail entries,
each with 512 dimensions.
During policy fine-tuning, LexiconVLA conditions execution
on the task instruction, the current subtask instruction,
and retrieved action codes.
Code retrieval uses observations available at the subtask
boundary, without access to future observations.

\paragraph{BridgeVLA Fine-tuning.}
All BridgeVLA comparison arms initialize from the official
RLBench checkpoint. Training starts with 100 demonstrations for each seen task. The base policy, subtask-only control, and full LexiconVLA
use the same demonstrations, sample order, and optimization
budget. PaliGemma remains frozen; the action head and, where present,
the code injector are trainable.
Table~\ref{tab:bridge-training} summarizes the configuration.

\begin{table}[htbp]
\centering
\small
\caption{Fine-tuning configuration for the BridgeVLA
comparison arms. These settings are specific to BridgeVLA.}
\label{tab:bridge-training}
\renewcommand{\arraystretch}{1.1}
\begin{tabular}{@{}ll@{}}
\toprule
\textbf{Setting} & \textbf{Value} \\
\midrule
Optimizer & AdamW \\
Learning rate & $8\times10^{-5}$ \\
Batch size & 8 \\
Training epochs & 3 \\
Optimizer steps & 17,904 \\
Training seed & 20260914 \\
Frozen modules & PaliGemma, codebooks, adapter \\
Trainable modules & Action head and optional code injector \\
\bottomrule
\end{tabular}
\end{table}

\paragraph{Test-Time Execution.}
Each task is evaluated on 25 episodes initialized from test
demonstrations, with identical initial scenes across comparison
arms.
Each episode permits at most 25 keyframe actions.
LexiconVLA uses an online planner to schedule subtasks.
At each subtask entry, the adapter retrieves codes from
the current observation and subtask instruction; these codes
are reused within the subtask.
All parameters remain fixed throughout evaluation.
An episode succeeds only when RLBench's task-level success
condition is satisfied.
Timeouts and execution failures count as unsuccessful episodes
and remain in the denominator. Unseen, seen, and overall success rates average per-task success
rates with equal weight across tasks, and we report their mean and
standard deviation over three random seeds.

\paragraph{Policy-Held-Out Evaluation.}
LexiconVLA separates atomic-level pretraining from downstream policy training. We pretrain both the action codebooks and the scene-aware adapter on AtomAction, which contains atomic segments from 69 RLBench tasks, including the eight downstream-policy-held-out tasks. The codebooks learn trajectory-derived action representations, while the adapter learns to retrieve the corresponding codes from observations and atomic instructions. Both components are then frozen during downstream policy training, which uses demonstrations from the 18 policy-training tasks and excludes demonstrations from the eight held-out tasks.

This protocol evaluates whether pretrained atomic representations and their retrieval interface can support execution of tasks excluded from downstream policy training. Here, policy-held-out specifies the downstream training boundary; it does not imply that the corresponding task families are absent from atomic-level pretraining. The intended transfer is from pretrained segment-level knowledge to downstream task execution, with no parameter updates during evaluation. We report results under this pretraining-augmented protocol and distinguish it from an all-stage task-disjoint evaluation.

\paragraph{Online Subtask Generation.}
At evaluation time, the full-task instruction is provided by RLBench, while atomic subtask instructions are generated by the online VLM planner using the task instruction and current visual observations. For the eight downstream-policy-held-out tasks, the reference action sequence field is left empty, and the planner's training-cache phrasings exclude annotated decompositions from these tasks. Thus, the task-specific subtask sequence used for execution is generated online rather than supplied as an annotated target-task decomposition.

The planner uses the predefined action vocabulary and prompt conventions to express its proposed subtasks. The adapter then retrieves codes for each active subtask from the available observation and generated instruction. Atomic-level supervision used to pretrain the adapter is distinct from supplying a ground-truth task decomposition to the planner at evaluation time.

\paragraph{Compared Methods.}
\begin{itemize}
    \item \textbf{Simulation Baselines.} All simulation baselines are keyframe-based 3D manipulation policies developed for, and evaluated on, the multi-task RLBench setting. We apply LexiconVLA to five of them, covering three architectural families: PerAct, a Perceiver transformer over voxelized observations; RVT and RVT-2, multi-view transformers over virtual renderings of the scene point cloud, with RVT-2 adding coarse-to-fine refinement; and BridgeVLA and BridgeVLA++, which build on a pretrained VLM by projecting 3D inputs to multi-view images and predicting 2D heatmaps, with BridgeVLA++ adding memory augmentation. Each base policy and its LexiconVLA variant are evaluated on identical initial scenes. Act3D, a 3D feature-field transformer; 3D Diffuser Actor, a diffusion policy over 3D scene representations; and SAM2Act, a multi-view policy built on SAM2 visual features, serve as standalone references for absolute performance.

    \item \textbf{Generalist VLAs.} Generalist VLAs such as \(\pi_0\) and \(\pi_{0.5}\) are not included in the simulation comparison. Their public releases provide no RLBench checkpoints, and they predict continuous action chunks at control frequency, whereas the RLBench protocol followed by all compared methods predicts sparse keyframe poses executed by a motion planner. Including them would require re-designing their action interface and training them on RLBench under our own choices, so the results would reflect our adaptation rather than the released models. On the real robot, this mismatch does not arise: \(\pi_{0.5}\) can be fine-tuned with its standard pipeline on the same Franka Research~3 demonstrations as the other real-world methods. We therefore include it in the real-world experiments as a generalist VLA reference with a different architecture and action representation, complementing BridgeVLA, which serves as the common base policy in both settings.
\end{itemize}


\section{Real-World Setup and Evaluation Protocol}
\label{app:real-setup}


\begin{figure*}[t]
    \centering
    \includegraphics[width=\linewidth]
    {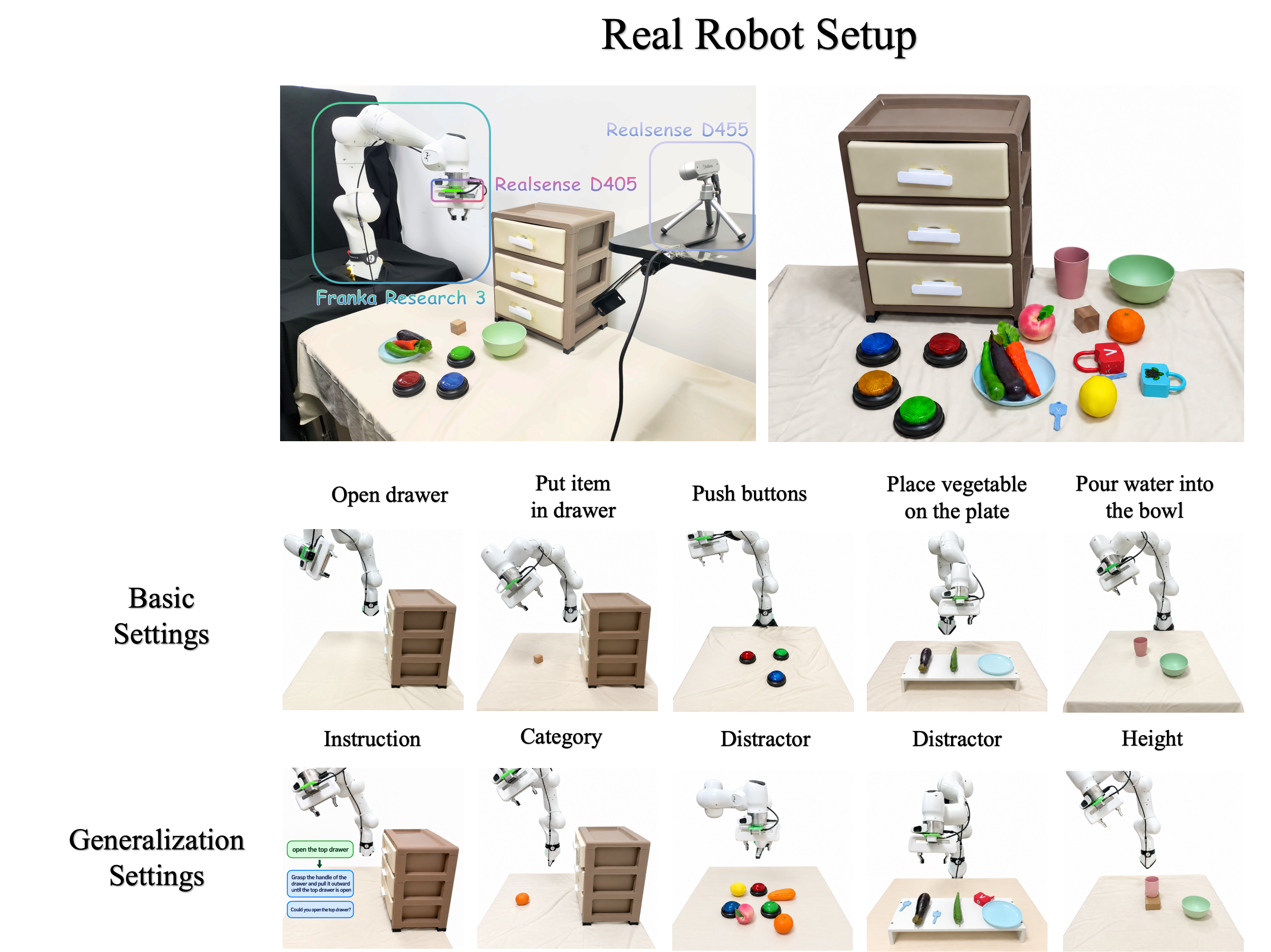}
    \caption{
    Real-robot platform and experimental settings.
    Top: the robot setup and all objects used in the experiments.
    Middle, from left to right: open drawer, put item in drawer,
    push buttons, place items on a plate, and pour water into a bowl.
    Bottom: the corresponding generalization settings involving
    instruction, category, distractor,
    and height.
    }
    \label{fig:real_world_setup}
\end{figure*}

\paragraph{Platform and Demonstrations.}
The real-world experimental setup is illustrated in Fig.~\ref{fig:real_world_setup}. All experiments are conducted on a 7-DoF Franka Research~3 (FR3) robotic arm equipped with a parallel-jaw gripper. Visual observations are captured from two complementary viewpoints: a RealSense D455 camera provides a fixed third-person view of the workspace, while a RealSense D405 camera is mounted on the robot wrist to capture close-range interaction details. To improve visual clarity and better emphasize the robot, manipulated objects, and task-relevant regions, we remove irrelevant background content and retain only the principal foreground elements in the presented figures. This post-processing is applied solely for visualization and does not alter the visual observations used by the policy during training or evaluation. We collect 30 demonstrations per task, covering all task variants in Table~\ref{tab:real-tasks}. The Standard setting uses the objects, scene layout, and instructions of the demonstrations.

\paragraph{Compared Methods and Training.}
All methods are fine-tuned on the same demonstrations. BridgeVLA and LexiconVLA (BridgeVLA) are initialized from the same pretrained checkpoint, and $\pi_{0.5}$ is fine-tuned with its standard pipeline from its released base checkpoint; Appendix~\ref{app:sim-setup} explains why $\pi_{0.5}$ is included only in real-world experiments. LexiconVLA keeps the codebooks learned on AtomAction frozen and uses the same planner and scheduling skills as in simulation (Appendix~\ref{app:lexiconvla-sch}). No parameters are updated at test time in either setting.

\paragraph{Standard Tasks.}
Table~\ref{tab:real-tasks} lists the five tasks and their success criteria. Fig.~\ref{fig:real_world_setup} shows the real-world scenes for the five standard tasks. Open drawer, put item in drawer, and push buttons have counterparts among the seen RLBench tasks (Sim), whereas pour water into the bowl and place vegetables on the plate have none (Real-only). Push buttons and place vegetables on the plate are treated as long-horizon because their instructions specify multiple targets that must be handled in sequence, so a failure at any target fails the trial.

\paragraph{Real-World Adaptation.}
We use real-world pouring to examine adaptation beyond the original atomic annotation vocabulary. Pouring is not an additional AtomAction label, and no codebook entries are added for this experiment. Instead, we retain the simulation-pretrained codebooks and fine-tune the downstream LexiconVLA policy using 30 real-world demonstrations of the pouring task. This setting evaluates whether a fixed pretrained atomic vocabulary can remain useful when the downstream policy learns a new target behavior.

LexiconVLA succeeds in 4 of 10 standard pouring trials, compared with 2 of 10 for BridgeVLA and 0 of 10 for $\pi_{0.5}$ under the reported training protocols. This result supports adaptation without expanding the lexicon at the evaluated budget of 30 demonstrations.


\begin{table*}[t]
\centering
\small
\caption{\textbf{Real-world tasks, variants, and success criteria.} Groups follow Table~\ref{tab:sim2real_generalization}.}
\label{tab:real-tasks}
\renewcommand{\arraystretch}{1.15}
\begin{tabularx}{\linewidth}{@{}>{\raggedright\arraybackslash}p{0.19\linewidth} >{\raggedright\arraybackslash}p{0.16\linewidth} >{\raggedright\arraybackslash}p{0.25\linewidth} >{\raggedright\arraybackslash}X@{}}
\toprule
\textbf{Task} & \textbf{Group} & \textbf{Variants} & \textbf{Success criterion} \\
\midrule
Open drawer & Sim-Short & Top / middle drawer & The specified drawer is pulled open. \\
Put item in drawer & Sim-Short & Top / middle drawer & The specified drawer is opened and the block rests inside it. \\
Push buttons & Sim-Long & Red; red$\to$blue; red$\to$blue$\to$green & All specified buttons are pressed in the given order. \\
Pour water into the bowl & Real-only-Short & Single setup & The cup is grasped, carried to the bowl, and tilted to pour into it without being knocked over. \\
Place vegetables on the plate & Real-only-Long & Pepper + eggplant; pepper + carrot; eggplant + carrot & Both items rest on the plate at the end of the trial. \\
\bottomrule
\end{tabularx}
\end{table*}

\paragraph{Generalization Settings.}
Each perturbation is applied to the task listed in Table~\ref{tab:sim2real_generalization} and is absent from the demonstrations. Figure~\ref{fig:real_world_setup} shows the real-world scenes under the generalization settings.
\begin{itemize}
    \item \textbf{Distractor} (push buttons; place vegetables on the plate). One or two distractors are placed near the targets, including objects of the target category (an additional yellow button; additional fruits such as an orange and a peach) and unrelated objects (fruits among the buttons; toys near the plate). This setting tests whether execution remains bound to the instructed targets.
    \item \textbf{Height} (pour water into the bowl). The manipulated objects are raised on supports of two different heights, shifting the grasping and pouring poses away from those demonstrated.
    \item \textbf{Category} (put item in drawer). The wooden block used in the demonstrations is replaced by objects of unseen categories (an orange and a lemon), which differ from the training item in shape, size, and surface.
    \item \textbf{Instruction} (open drawer). The training instruction is replaced by a detailed procedural command, ``Grasp the handle of the drawer and pull it outward until the top/middle drawer is open,'' and a conversational request, ``Could you open the drawer?''
\end{itemize}

\paragraph{Evaluation Protocol.}
Each method is evaluated over 10 trials per task and setting. When a task has several variants, trials alternate among them, and all methods follow the same variant sequence. A trial succeeds only if the entire instruction is completed; partial completion, such as pressing only part of a button sequence or leaving one item off the plate, counts as failure. Trials terminated by collisions, joint or workspace limits, or controller faults also count as failures and remain in the denominator. The human-disturbance rollouts in Figure~\ref{fig:real_world_demos} are qualitative and are not included in Table~\ref{tab:sim2real_generalization}.


\section{Detailed Real-World Analysis}
\label{app:real-analysis}


\begin{table}[t]
\centering
\caption{\textbf{Aggregated real-world success rates (\%).} Group averages of Table~\ref{tab:sim2real_generalization}. Each task appears once in each setting, so both All columns cover the same five tasks.}
\label{tab:real-aggregate}
\footnotesize
\setlength{\tabcolsep}{3pt}
\begin{tabular}{@{}lcccc@{}}
\toprule
\multirow{2}{*}{\textbf{Method}} & \multicolumn{3}{c}{\textbf{Standard}} & \makecell{\textbf{Generalization}} \\
\cmidrule(lr){2-4} \cmidrule(lr){5-5}
 & \makecell{Sim\\(3)} & \makecell{Real-only\\(2)} & \makecell{All\\(5)} & \makecell{All\\(5)} \\
\midrule
BridgeVLA & 36.7 & 10.0 & 26.0 & 16.0 \\
$\pi_{0.5}$ & 46.7 & 0.0 & 28.0 & 12.0 \\
\rowcolor{gray!10} LexiconVLA (Ours) & \textbf{73.3} & \textbf{45.0} & \textbf{62.0} & \textbf{56.0} \\
\bottomrule
\end{tabular}
\end{table}

\paragraph{Standard Setting.}
As summarized in Table~\ref{tab:real-aggregate}, LexiconVLA outperforms both baselines on tasks with and without simulation counterparts, and its advantage grows with the number of chained atomic actions. On open drawer, LexiconVLA succeeds in every trial, compared with 50.0\% for BridgeVLA and 60.0\% for $\pi_{0.5}$. Put item in drawer appends grasping, transferring, and placing an item to the same opening interaction. Here both baselines fail in all trials, often already at the opening stage by missing the handle or pulling a different drawer. An interaction a policy performs in one task is thus not reliably invoked within another, in line with our diagnostic probe (Appendix~\ref{app:diagnostic}). Most remaining failures of LexiconVLA occur after the drawer is opened, while grasping or transferring the item. On place vegetables on the plate, which repeats pick-and-place for two items, several LexiconVLA failures are partial completions in which the second item falls off the plate; these count as failures under our criterion. On pour water into the bowl, BridgeVLA frequently knocks the cup over or reaches joint limits, and $\pi_{0.5}$ fails in all trials, in some cases executing the pouring motion without a stable grasp on the cup. Push buttons is the only task on which a baseline matches LexiconVLA ($\pi_{0.5}$, 80.0\%), and this parity disappears under distractors.

\paragraph{Generalization Setting.}
Figure~\ref{fig:generalization} illustrates the four generalization settings: Instruction, Category, Height, and Distractor. Because each task is perturbed exactly once, the Generalization averages are directly comparable with the Standard ones: LexiconVLA retains about 90\% of its Standard success rate (56.0\% vs.\ 62.0\%), compared with about 62\% for BridgeVLA and 43\% for $\pi_{0.5}$.








\begin{figure*}[t]
    \centering

    \includegraphics[width=0.65\linewidth]{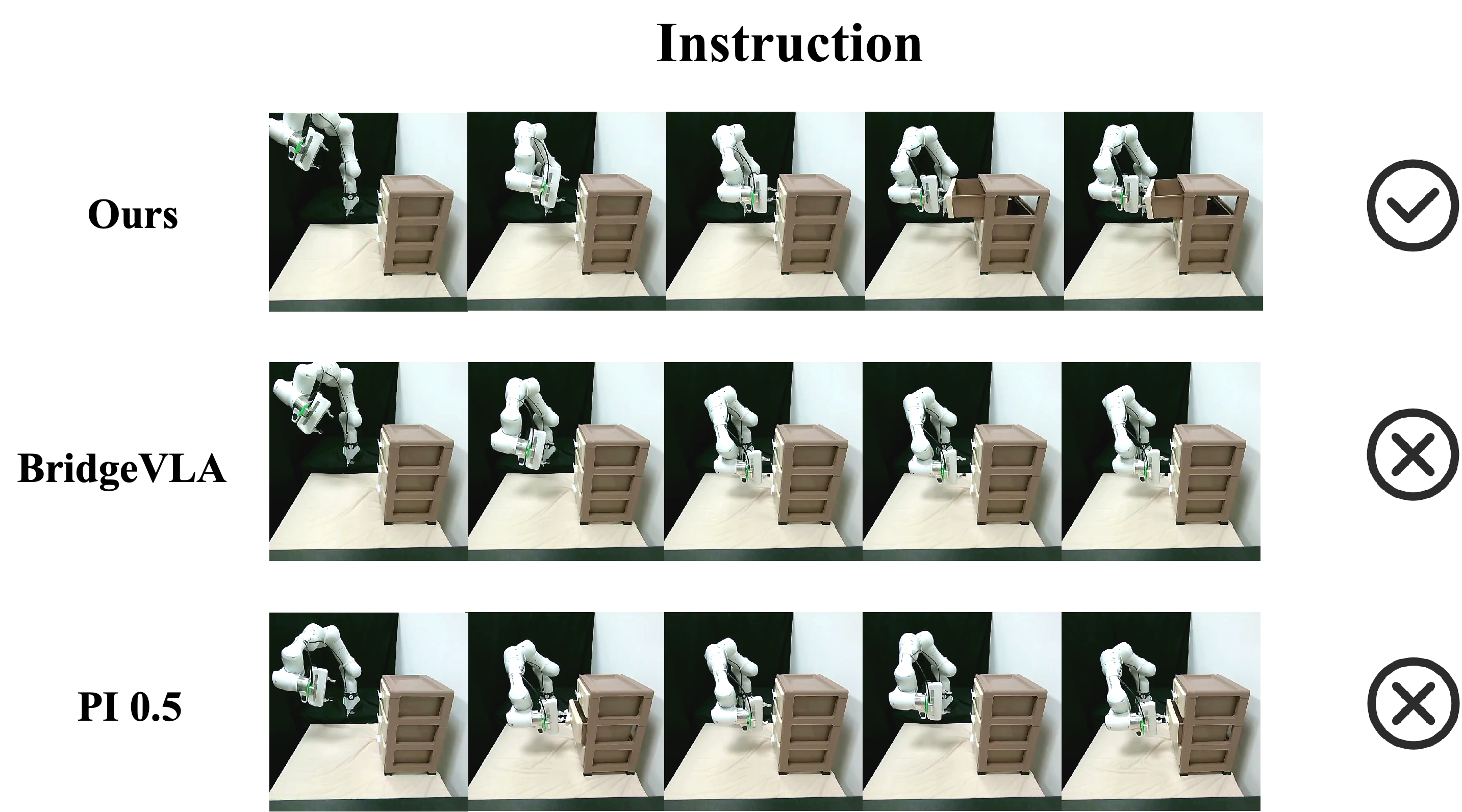}

    \vspace{3pt}
    \includegraphics[width=0.65\linewidth]{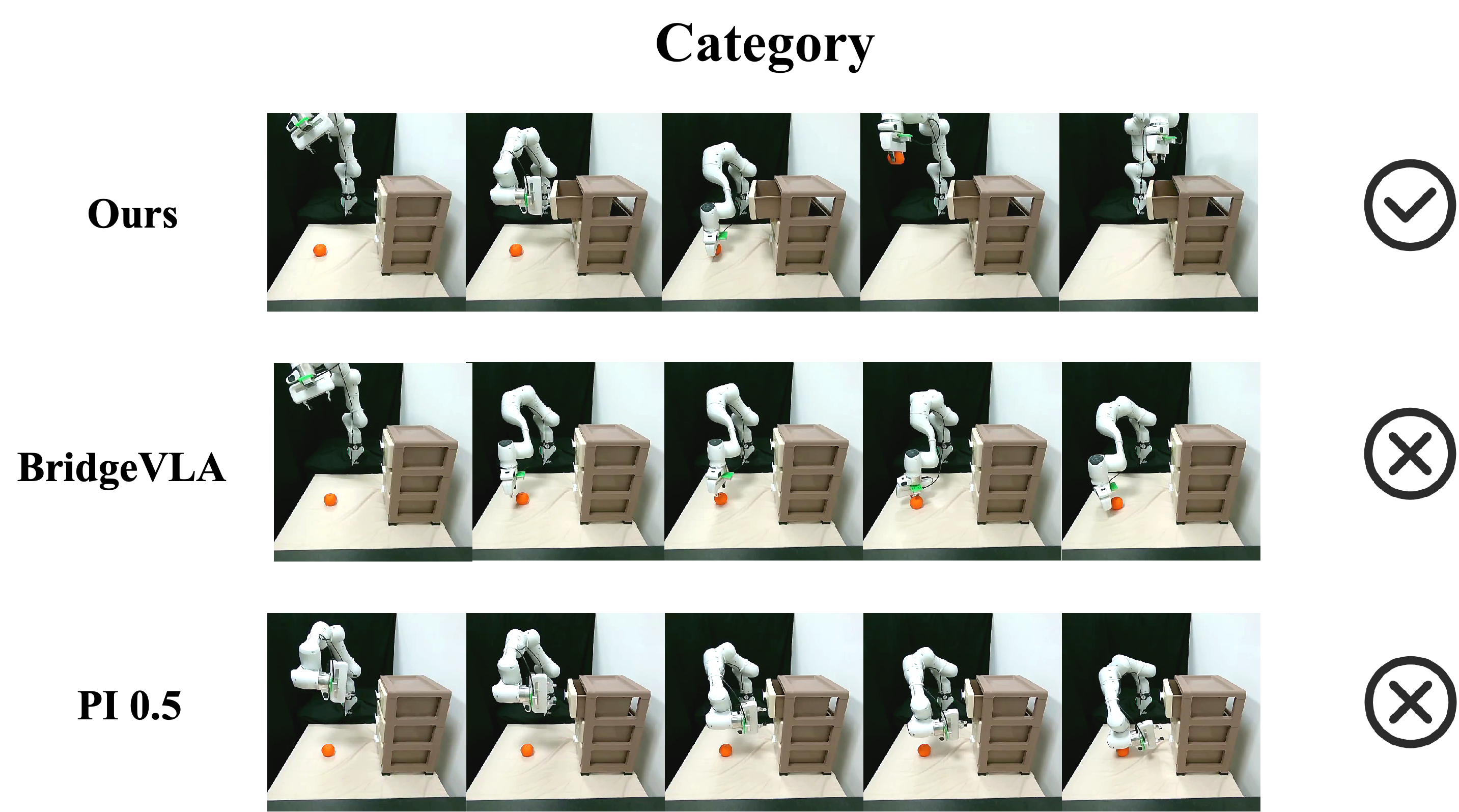}

    \vspace{3pt}
    \includegraphics[width=0.65\linewidth]{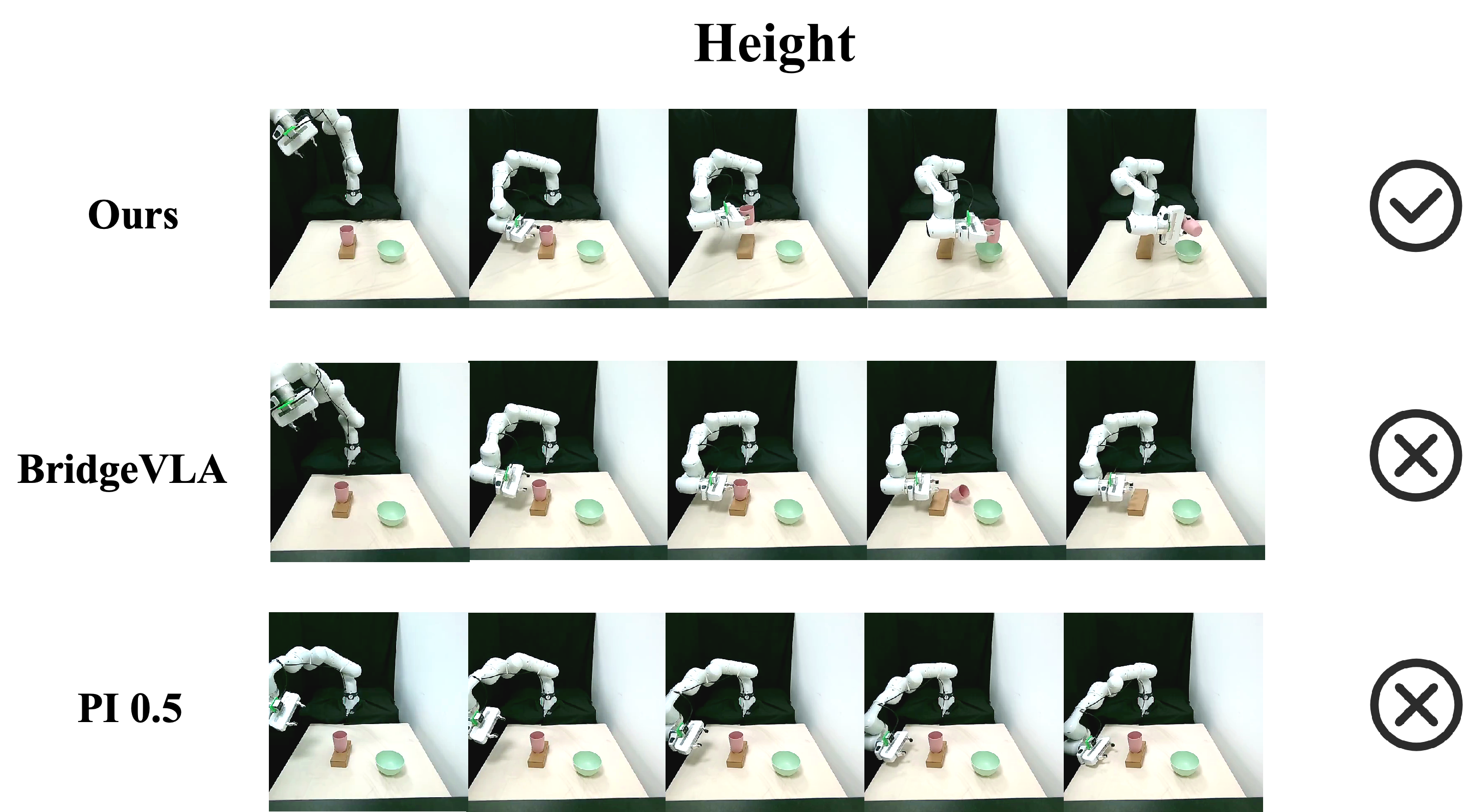}

    \caption{
\textbf{Real-world rollouts under generalization settings and human disturbance.}
The first four rows show examples in which LexiconVLA succeeds under the Instruction, Category, Height, and Distractor settings, while BridgeVLA and $\pi_{0.5}$ fail.
\textbf{Instruction:} BridgeVLA and $\pi_{0.5}$ both open the wrong drawer.
\textbf{Category:} Both baselines fail to grasp the object correctly.
\textbf{Height:} BridgeVLA fails to maintain a stable grasp on the cup at the changed height, whereas $\pi_{0.5}$ does not attempt to grasp it.
\textbf{Distractor:} Both baselines fail to grasp the correct object.
The fifth row shows a \textbf{human-disturbance} example in which LexiconVLA succeeds, while BridgeVLA fails by grasping the wrong drawer.
}
\label{fig:generalization}
\end{figure*}

\begin{figure*}[t]
    \ContinuedFloat
    \centering

    \includegraphics[width=0.65\linewidth]{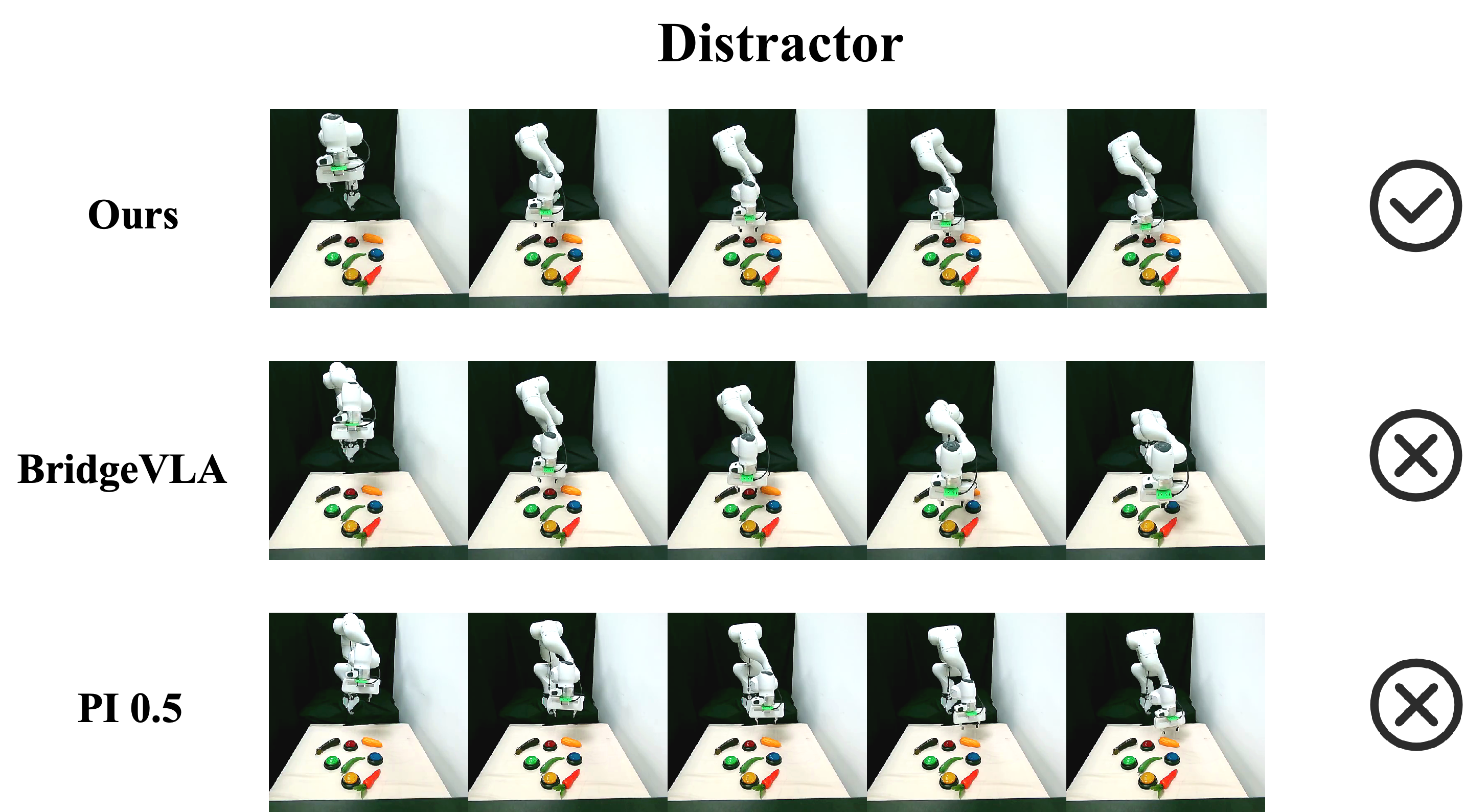}

    \vspace{3pt}
    \includegraphics[width=0.65\linewidth]{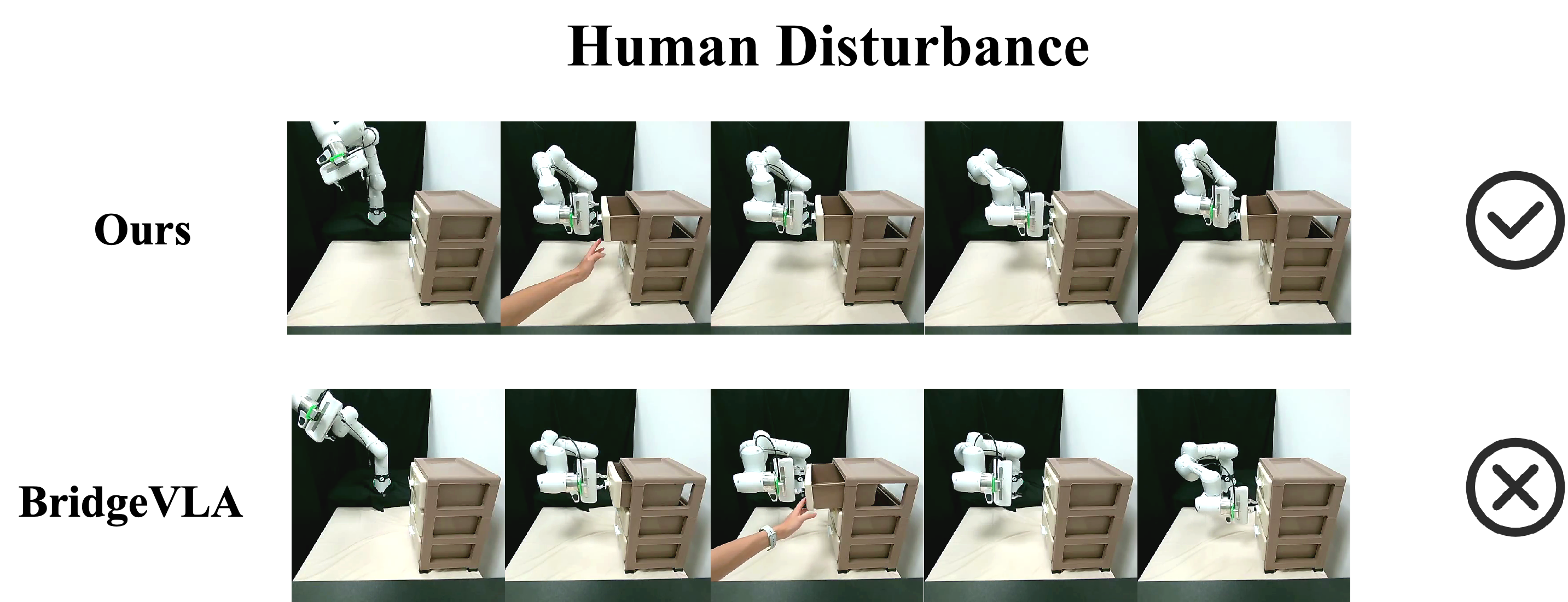}

    \caption{
        \textbf{Real-world rollouts under generalization settings and human disturbance (continued).}
    }
\end{figure*}

\begin{itemize}
    \item \textbf{Instruction.} Paraphrases lower $\pi_{0.5}$ from 60.0\% to 20.0\% and BridgeVLA from 50.0\% to 40.0\% on open drawer, whereas LexiconVLA keeps 80.0\%. Several $\pi_{0.5}$ failures pull a drawer other than the instructed one. For LexiconVLA, the planner maps both paraphrases onto the same atomic subtasks, so the subtask instructions and retrieved codes that condition execution remain largely unchanged.
    \item \textbf{Category.} Replacing the wooden block with an orange or a lemon leaves LexiconVLA at 40.0\%, its Standard success on this task, with failures arising mainly from grasping the round, smooth objects. As both baselines already fail in the Standard setting, this condition mainly shows that the success of LexiconVLA is not tied to the demonstrated object, consistent with codes that encode the interaction rather than object identity.
    \item \textbf{Height.} LexiconVLA reaches 80.0\% on raised objects, whereas both baselines fail in all trials. Its success exceeds that in the Standard setting partly because the raised placement reduces reachability failures at table height; the relevant comparison is therefore across methods under the same condition. Baseline failures, such as closing the gripper before reaching the cup, colliding with the support, or pouring in place, suggest motions tied to the demonstrated heights, whereas LexiconVLA retrieves codes from the current observation at each subtask boundary.
    \item \textbf{Distractor.} Distractors are the most challenging perturbation for LexiconVLA, lowering its success from 80.0\% to 60.0\% on push buttons and from 50.0\% to 20.0\% on place vegetables on the plate. It nevertheless leads on both tasks, while $\pi_{0.5}$ drops from 80.0\% to 40.0\% on push buttons. Failures of all methods mainly involve acting on a distractor, such as pressing a fruit placed among the buttons or confusing the pepper with a similar-looking item. The codes specify which interaction to perform but not which object to act on, so target binding under same-category distractors still relies on the planner and base policy; object-centric grounding is a natural extension.
\end{itemize}

\paragraph{Failure Recovery.}
Figures~\ref{fig:real_world_demos} and~\ref{fig:generalization} show two forms of recovery. Without external intervention, a failed grasp in put the ball in the hoop is detected by the planner, which issues RETRY to re-execute the failed subtask. The policy uses updated observations with the existing code conditioning during recovery. Under human disturbance in put item into the drawer, the planner likewise returns to the grasp subtask and completes the task, whereas the $\pi_{0.5}$ rollout fails under the same disturbance. In contrast, baselines can stall after a failed interaction; for example, $\pi_{0.5}$ repeatedly closes its gripper near a missed drawer handle without re-approaching it. In another human-disturbance scenario, an already opened drawer is closed again; the planner reopens the correct drawer, whereas BridgeVLA attempts to open the wrong one. Recovery does not guarantee completion, however, as subsequent grasps or transfers can still fail.

\section{Atomic-Action Standard}
\label{app:action-standard}

This appendix specifies the operational meaning of each released action label. Each entry pairs a normalized definition with the start and terminal frames of the reference example provided in the annotation workbook. The images illustrate representative state changes; the label definitions, rather than scene appearance alone, determine annotation.

\actionstandard{Approach}{The end effector moves toward a target object or contact site without making physical contact. Its orientation may change during the motion, and the phase ends at a preparatory pose for the next interaction.}{Beat the buzz}{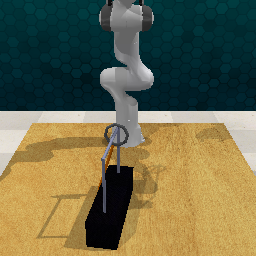}{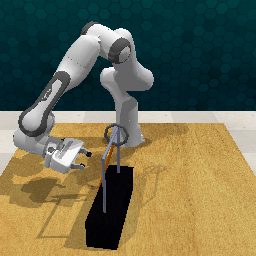}

\actionstandard{Grasp}{From a pre-grasp pose, the end effector advances until contact and closes the gripper around the target object. The phase establishes a stable hold through friction or geometric constraint.}{Pick and lift}{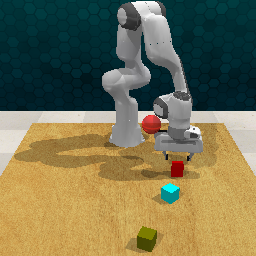}{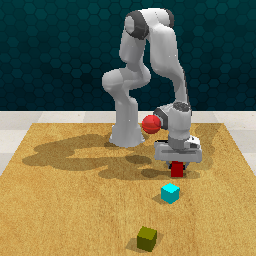}

\actionstandard{Lift}{While maintaining a stable grasp, the end effector moves the target object upward until it is fully separated from its initial support surface.}{Meat off grill}{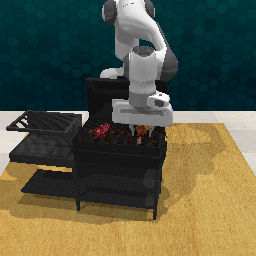}{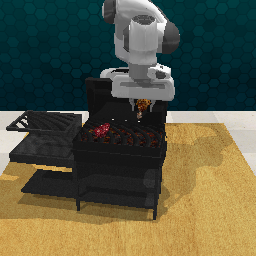}

\actionstandard{Transfer}{While maintaining a stable grasp after lift, the end effector carries the object from its current location toward a pre-placement pose near the target region.}{Put item in drawer}{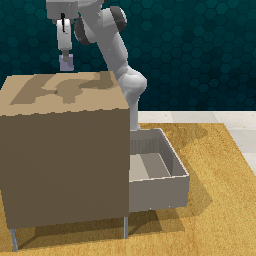}{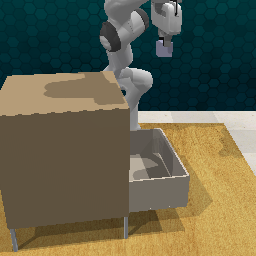}

\actionstandard{Rotate}{While maintaining a stable grasp, the end effector changes the object's orientation through angular motion about an axis. The object's overall position remains approximately fixed unless a more specific interaction label applies.}{Change clock}{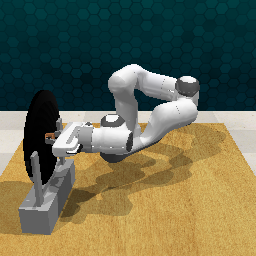}{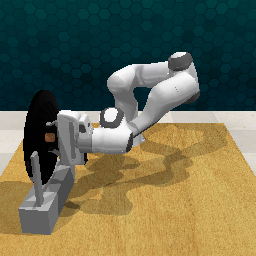}

\actionstandard{Revolve-out}{The end effector applies tangential motion to a component constrained by a fixed axis, such as a hinge, causing it to rotate outward and away from its closed or initial configuration.}{Open window}{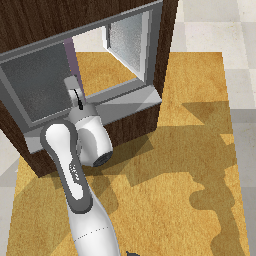}{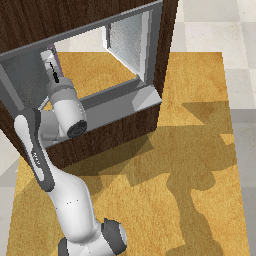}

\actionstandard{Revolve-in}{The end effector applies tangential motion to a component constrained by a fixed axis, causing it to rotate inward toward its closed configuration. Contact may be maintained without a grasp when pushing is sufficient.}{Close microwave}{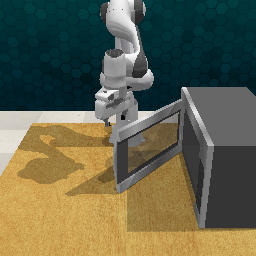}{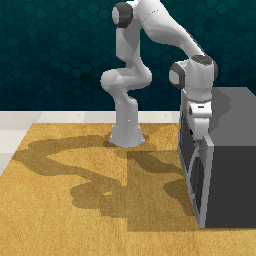}

\actionstandard{Push}{After contact is established, the end effector applies force away from itself along a constrained direction, producing translational or mechanically constrained displacement of the target object or mechanism.}{Hit ball with queue}{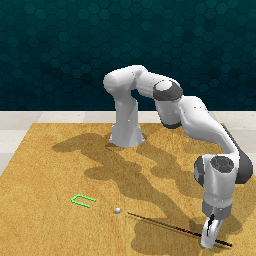}{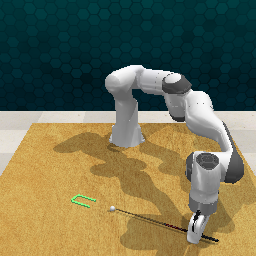}

\actionstandard{Flip-close}{The end effector acts on a component that rotates about a predominantly horizontal hinge or pivot, moving it downward from an open state until it reaches the closed configuration.}{Close laptop lid}{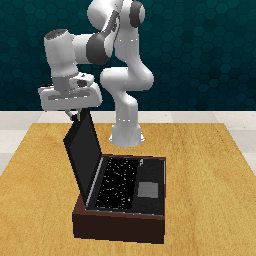}{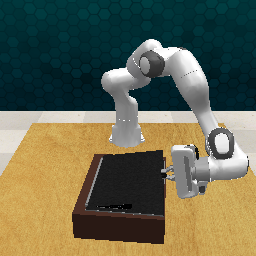}

\actionstandard{Place}{From a pre-placement pose, the end effector aligns and lowers a held object onto a target support or into a target region, establishing the intended final support relationship.}{Place shape}{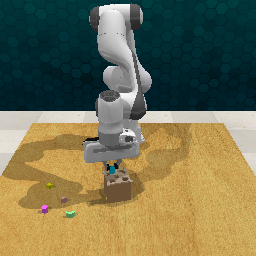}{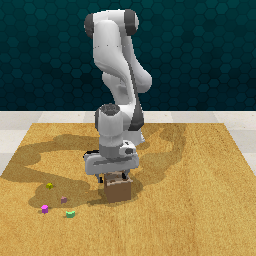}

\actionstandard{Press}{After establishing contact with a control or surface, the end effector applies a short displacement and controlled force along the contact normal, typically until a mechanical travel limit is reached.}{Get ice from fridge}{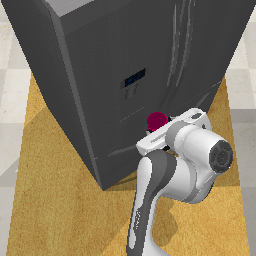}{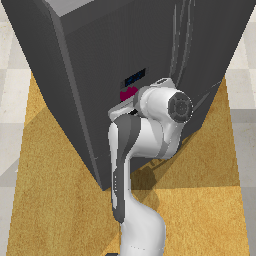}

\actionstandard{Hang}{The end effector aligns an object's hook, loop, hole, or mounting feature with a fixed support and seats it so that the object remains suspended by that support.}{Hang frame on hanger}{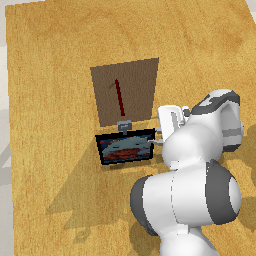}{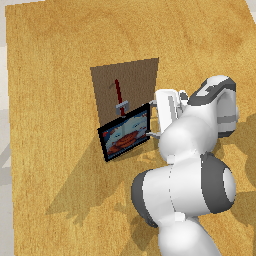}

\actionstandard{Insert}{The end effector aligns a held object with a compatible hole, slot, or interface and advances it along the insertion axis until the intended geometric engagement is established.}{Insert usb in computer}{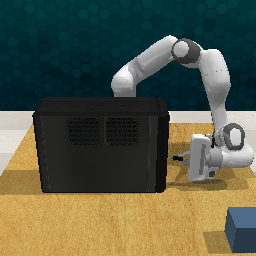}{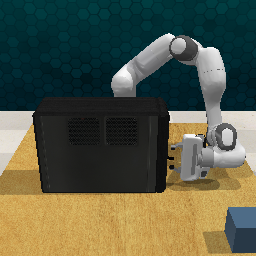}

\actionstandard{Flip-open}{The end effector maintains contact with a lid, cover, flap, or similar component and rotates it about a predominantly horizontal hinge or pivot from a closed state to an open state.}{Open grill}{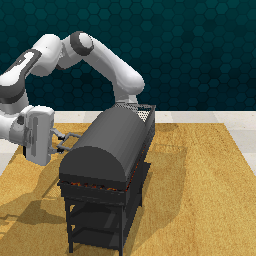}{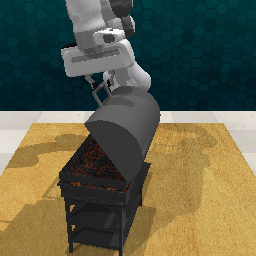}

\actionstandard{Pull}{While maintaining a stable grasp on an object or mechanism handle, the end effector moves toward itself along a constrained axis, opening, extending, or extracting the target.}{Open drawer}{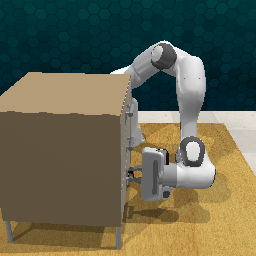}{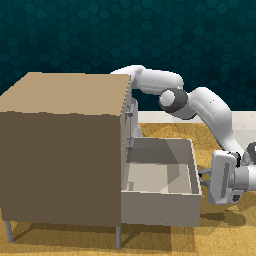}

\actionstandard{Slide}{The end effector maintains a grasp or firm contact while moving an object or sliding mechanism along its constrained linear axis without lifting it from that constraint.}{Reach and drag}{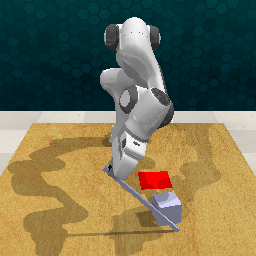}{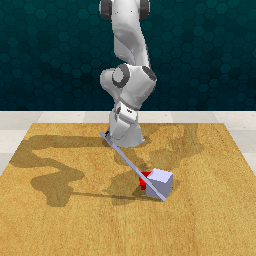}

\actionstandard{Pose-adjust}{The end effector performs a local correction to its pose or to the pose of a grasped object so that the following interaction can proceed from a better-aligned configuration.}{Screw nail}{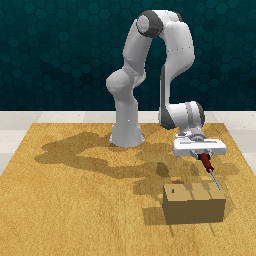}{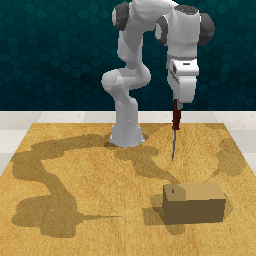}

\actionstandard{Wipe}{The end effector maintains continuous surface contact, directly or through a held tool, while sweeping across a region to produce a covering or cleaning motion.}{Wipe desk}{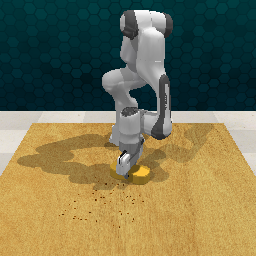}{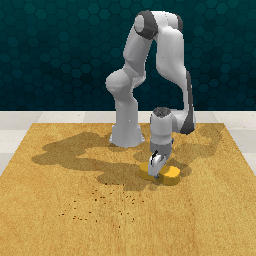}


\section{Limitations and Future Work}
\label{sec:limitations-future-work}
LexiconVLA investigates how atomic-level pretraining can support downstream-policy transfer within the data coverage specified above. Evaluating task families excluded from both atomic pretraining and downstream training would further characterize the generalization of the learned lexicon. The current vocabulary is learned from a finite set of simulated interaction phases, and its coverage of substantially different contact dynamics and embodiments remains to be established. Incorporating more diverse interaction data and incrementally extending the lexicon are natural directions for broadening this coverage. Online VLM planning supports feedback-driven execution, but introduces additional inference calls and makes execution sensitive to subtask decomposition and monitoring decisions. More efficient scheduling, uncertainty-aware monitoring, and controlled recovery evaluations could improve and better characterize the reliability of this interface. Our real-world experiments cover one robot platform and a limited set of tasks and trials; larger evaluations across embodiments and repeated runs would strengthen estimates of transfer performance. In particular, pouring demonstrates adaptation with an unchanged vocabulary using 30 demonstrations, while learning curves across demonstration budgets are needed to assess sample efficiency. These directions build on the central contribution of LexiconVLA: a trajectory-derived global/detail action lexicon, grounded through visual state-transition supervision and made accessible to a shared execution policy. The integration across multiple policy backbones and the real-world adaptation results support the practical value of this interface under the evaluated conditions, providing a foundation for studying how reusable atomic knowledge can be acquired, transferred, and expanded across manipulation tasks. Although LexiconVLA achieves positive average gains on policy-held-out tasks across all five evaluated backbones, the magnitude of improvement varies considerably. Gains are relatively modest for PerAct, RVT, and RVT-2, while substantially larger improvements are observed with BridgeVLA and BridgeVLA++. This variation suggests that the benefit of atomic-code conditioning depends in part on the underlying policy's ability to utilize subtask-level guidance. Nevertheless, the positive average gains across different policy architectures provide evidence of the applicability of the proposed interface under the evaluated settings, rather than improvements restricted to a single backbone. More extensive controlled evaluations across policy architectures, training configurations, and task distributions are needed to characterize this dependency and establish when reusable atomic representations are most effective.


\section{Prompt Templates}
\label{app:prompt}
\label{app:planner-prompts}
This appendix collects the prompt templates used in this work, all abbreviated
for presentation. Table~\ref{tab:phase_description_prompt} presents the phase-level language generation prompt, which produces the eight English descriptions attached to every atomic-action segment of the dataset. The remaining two templates drive the online closed loop execution, which queries the VLM planner with two prompts: PLAN(Table~\ref{tab:plan_prompt}) decomposes the remaining task into atomic-action subtasks, and MONITOR (Table~\ref{tab:monitor_prompt}) selects a scheduling
skill for the active subtask. The scheduling skills and the call, retry, and extension limits referenced below follow Appendix~\ref{app:lexiconvla-sch}.

\begin{table*}[t]
\centering
\caption{
Prompt template and example for phase-level language generation.
}
\begin{minipage}{0.96\linewidth}
\small
\setlength{\parindent}{0pt}
\setlength{\parskip}{2pt}

\hrule height 0.8pt
\vspace{5pt}

\textbf{Role:}

You are annotating phases for an RLBench robot manipulation dataset.
Generate reusable English descriptions for each exact phase.
You must be accurate, diverse in wording and form, and strictly output
valid JSON.

\vspace{4pt}
\textbf{Input:}

The input is a JSON object containing:

\begin{enumerate}
    \setlength{\itemsep}{0pt}
    \setlength{\parsep}{0pt}
    \setlength{\topsep}{1pt}
    \setlength{\partopsep}{0pt}

    \item Task and variation information, including task,
    variation index, and variation descriptions.

    \item The complete phase label sequence and one item for every
    phase. Each item contains phase index, action label,
    previous action label, next action label,
    repeated action label, and same action phase indices.

    \item The descriptions per phase value and the list of
    required styles.
\end{enumerate}

\textbf{Example input:}
task = open drawer; variation = 1; phase sequence =
[approach, grasp, pull]; descriptions per phase = 8.

\vspace{4pt}
\textbf{Task:}

Generate phase-level English descriptions for every item in the input JSON.

\vspace{4pt}
\textbf{Requirements:}

\setlength{\columnsep}{18pt}
\begin{multicols}{2}
\begin{enumerate}
    \setlength{\itemsep}{1pt}
    \setlength{\parsep}{0pt}
    \setlength{\topsep}{1pt}
    \setlength{\partopsep}{0pt}

    \item Return JSON only, with the top-level key
    phase descriptions.

    \item phase descriptions must contain one entry for every input
    item, ordered by phase index.

    \item Each entry must contain phase index, action label, and
    descriptions.

    \item descriptions must contain exactly one object for each
    required style.

    \item Each description must stay correct for this exact phase only,
    not the whole task.

    \item Keep the writing varied in wording, sentence form, and
    abstraction level.

    \item Mention the action, manipulated object, local target, or
    immediate subgoal when supported.

    \item Do not invent unsupported details such as color, material,
    left/right, exact spatial layout, or hidden states.

    \item If the same action label appears in multiple phases,
    distinguish the target phase using its phase index and neighboring
    labels.

    \item All text must be English.
\end{enumerate}
\end{multicols}

\textbf{Required styles:}

short, clear, object focused, goal focused, detailed, abstract,
natural, and instructional.

\vspace{4pt}
\textbf{Output Format:}

The response contains the top-level key phase descriptions. Each
phase entry contains phase index, action label, and eight
\{style, text\} description objects.

\vspace{5pt}
\hrule height 0.6pt
\vspace{5pt}

\textbf{Example Output:}

\textbf{Task:} open drawer
\hfill
\textbf{Phase:} 2
\hfill
\textbf{Action:} pull

\vspace{4pt}

\begin{minipage}[t]{0.485\linewidth}
\raggedright

\textbf{Short:}
Pull the middle drawer open.

\textbf{Clear:}
Draw the middle drawer outward using the grasped handle.

\textbf{Object-focused:}
The middle drawer is pulled outward by its handle.

\textbf{Goal-focused:}
Open the middle drawer by pulling on the handle.

\end{minipage}
\hfill
\begin{minipage}[t]{0.485\linewidth}
\raggedright

\textbf{Detailed:}
With the handle already grasped, move the gripper outward to slide the
middle drawer open.

\textbf{Abstract:}
Convert the secured grip into an opening motion.

\textbf{Natural:}
Pull back on the handle to open the middle drawer.

\textbf{Instructional:}
Pull the grasped middle drawer handle outward until the drawer opens.

\end{minipage}

\vspace{5pt}
\hrule height 0.8pt

\end{minipage}

\label{tab:phase_description_prompt}
\end{table*}

\begin{table*}[t]
\centering
\caption{
Abbreviated PLAN prompt used to decompose a task instruction into
atomic-action subtasks.
}
\begin{minipage}{0.96\linewidth}
\small
\setlength{\parindent}{0pt}
\setlength{\parskip}{2pt}

\hrule height 0.8pt
\vspace{5pt}

\textbf{Role:}

You are a task planner for a robot arm working on a tabletop. Given the
current front and wrist views, decompose the remaining task into an ordered
sequence of atomic actions. Return a single JSON object only.

\vspace{4pt}
\textbf{Constraints:}

\begin{enumerate}
    \setlength{\itemsep}{1pt}
    \setlength{\parsep}{0pt}
    \setlength{\topsep}{1pt}
    \setlength{\partopsep}{0pt}

    \item Choose action labels from the predefined vocabulary: approach,
    grasp, (remaining atomic action labels omitted). Use pose-adjust only
    when no other action applies.

    \item Each instruction describes one step: lowercase, no trailing
    punctuation, 3--8 words, imperative mood. Preserve distinguishing
    colours, ordinals, and spatial references.

    \item Use the prescribed initial verbs and do not substitute synonyms:
    approach $\rightarrow$ move toward, approach;
    grasp $\rightarrow$ grasp;
    (remaining action-to-verb mappings omitted).

    \item Plan only unfinished work, typically 3--8 steps. Treat the
    reference sequence as a soft prior, and obey any supplied repetition
    count exactly.
\end{enumerate}

\vspace{4pt}
\textbf{Action definitions:}

approach: Move toward the target without contact, stopping at a preparatory
pose. grasp: Establish contact and close the gripper to secure the object.
(Remaining action definitions omitted.)

\vspace{4pt}
\textbf{Instruction selection:} prefer A over B over C.

\begin{enumerate}
    \setlength{\itemsep}{1pt}
    \setlength{\parsep}{0pt}
    \setlength{\topsep}{1pt}
    \setlength{\partopsep}{0pt}
    \renewcommand{\labelenumi}{\Alph{enumi}.}

    \item Reuse a listed instruction via its phrasing identifier.

    \item Reuse its structure and substitute only the differing object or
    variation words.

    \item If no entry applies, imitate the listed grammar, length, and
    vocabulary.
\end{enumerate}

\vspace{4pt}
\textbf{Output format:}

\texttt{\{"plan": [\textless step\textgreater, ...]\}}, where each step takes
one of the following forms:

\texttt{\{"action": "grasp", "phrasing\_id": \textless id\textgreater\}}

\texttt{\{"action": "grasp", "phrasing\_id": \textless id\textgreater,
"substitute": \{"\textless old\textgreater": "\textless
new\textgreater"\}\}}

\texttt{\{"action": "\textless action\textgreater", "instruction":
"\textless subtask instruction\textgreater"\}}

\vspace{5pt}
\hrule height 0.6pt
\vspace{5pt}

\textbf{User message:}

\texttt{task: \{task\_name\}}

\texttt{task instruction: \{task\_instruction\}}

\texttt{reference action sequence (soft prior): \{reference\_sequence\}}

\texttt{phrasings from the training cache:}

\texttt{[0] \{action\_0\}: \{instruction\_0\}}

\texttt{[1] \{action\_1\}: \{instruction\_1\}}

(Remaining candidate entries omitted.)

\texttt{\{repetition\_constraint\_if\_applicable\}}

\texttt{\{completed\_steps\_if\_extending\}}

\texttt{\{attempt\_history\_if\_enabled\}}

\texttt{front view: \{front\_image\}} \hfill
\texttt{wrist view: \{wrist\_image\}}

\vspace{5pt}
\hrule height 0.8pt

\end{minipage}

\label{tab:plan_prompt}
\end{table*}

\begin{table*}[t]
\centering
\caption{
Abbreviated MONITOR prompt used to assess execution of the active subtask
and to select a scheduling skill.
}
\begin{minipage}{0.96\linewidth}
\small
\setlength{\parindent}{0pt}
\setlength{\parskip}{2pt}

\hrule height 0.8pt
\vspace{5pt}

\textbf{Role:}

You are a progress monitor for a robot arm executing a plan. Assess the
current images, the active step, the elapsed keyframes, and the gripper
state. Return a single JSON object only.

\vspace{4pt}
\textbf{Scheduling skills:}

\textbf{CONTINUE:} The active step is still in progress and remains
achievable.

\textbf{NEXT:} Its goal is visibly achieved; advance to the next step.

\textbf{RETRY:} A step failed. Provide an index to repeat an earlier step,
or omit it to retry the current step.

\textbf{REPLAN:} The environment changed and invalidated the plan;
regenerate it from the current scene, discarding progress.

\textbf{EXTEND:} Available on the final step when the task remains
unfinished. Append the remaining steps while preserving completed work, and
use EXTEND instead of NEXT at the end of the plan.

\vspace{4pt}
\textbf{Guidance:}

Judge completion from the images rather than from step counts. Do not replan
merely because the arm has not reached its target. Avoid indefinite CONTINUE
decisions and advance once completion is visible. Prefer NEXT or CONTINUE
when either applies.

\vspace{4pt}
\textbf{Output format:}

\texttt{\{"decision": "\textless skill\textgreater", "index": \textless
optional step index\textgreater, "reason": "\textless at most 12
words\textgreater"\}}

\vspace{5pt}
\hrule height 0.6pt
\vspace{5pt}

\textbf{User message:}

\texttt{task: \{task\_name\}} \hfill
\texttt{task instruction: \{task\_instruction\}}

\texttt{gripper: \{gripper\_state\}} \hfill
\texttt{keyframes spent on the current step: \{elapsed\_keyframes\}}

\texttt{plan:}

\texttt{[0] \{action\_0\}: \{instruction\_0\}}

\texttt{[1] \{action\_1\}: \{instruction\_1\}}

(Remaining steps omitted; the active step is marked with
\texttt{\textless-- CURRENT}.)

\texttt{\{gripper\_semantics\_note\_if\_applicable\}}

\texttt{\{final\_step\_note\_if\_applicable\}}

\texttt{\{progress\_warning\_if\_applicable\}}

\texttt{\{attempt\_history\_if\_enabled\}}

\texttt{front view: \{front\_image\}} \hfill
\texttt{wrist view: \{wrist\_image\}}

\vspace{5pt}
\hrule height 0.8pt

\end{minipage}

\label{tab:monitor_prompt}
\end{table*}

\section{Per-Task Simulation Results}
\label{app:per_task}
Tables~\ref{tab:sim-unseen} and~\ref{tab:sim-seen} list the per-task success rates underlying Table~\ref{tab:success_rates}. Each entry is the mean$_{\pm\text{std}}$ over three random seeds with 25 episodes per task and seed. The Avg.\ column averages the per-task means with equal weight and therefore reproduces the Unseen and Seen columns of Table~\ref{tab:success_rates} up to rounding; its std is computed across the three seed-level averages.

\begin{table*}[htbp!]
\centering
\caption{\textbf{Per-task success rate (\%) on the 8 unseen tasks.} These eight tasks are held out from downstream policy training. Mean$_{\pm\text{std}}$ over three random seeds. Best result per column in bold. Shaded rows are LexiconVLA variants.}
\label{tab:sim-unseen}
\setlength{\tabcolsep}{4pt}
\renewcommand{\arraystretch}{1.15}
\resizebox{\textwidth}{!}{%
\begin{tabular}{@{}lc|cccccccc@{}}
\toprule
\multirow{2}{*}{\textbf{Method}} & \multirow{2}{*}{\textbf{Avg.}} & \textbf{Basketball} & \textbf{Close} & \textbf{Knife on} & \textbf{Lamp} & \textbf{Open} & \textbf{Phone} & \textbf{Pick} & \textbf{Press} \\
 &  & \textbf{in Hoop} & \textbf{Drawer} & \textbf{Board} & \textbf{On} & \textbf{Jar} & \textbf{on Base} & \textbf{up Cup} & \textbf{Switch} \\
\midrule
Act3D & $15.7_{\pm2.5}$ & $1.3_{\pm2.3}$ & $68.0_{\pm4.0}$ & $1.3_{\pm2.3}$ & $0.0_{\pm0.0}$ & $0.0_{\pm0.0}$ & $0.0_{\pm0.0}$ & $18.7_{\pm4.6}$ & $36.0_{\pm10.6}$ \\
3D Diffuser Actor & $28.7_{\pm0.3}$ & $0.0_{\pm0.0}$ & $\mathbf{94.7}_{\pm4.6}$ & $5.3_{\pm2.3}$ & $0.0_{\pm0.0}$ & $0.0_{\pm0.0}$ & $2.7_{\pm2.3}$ & $42.7_{\pm2.3}$ & $\mathbf{84.0}_{\pm6.9}$ \\
SAM2Act & $15.5_{\pm1.5}$ & $0.0_{\pm0.0}$ & $42.7_{\pm2.3}$ & $0.0_{\pm0.0}$ & $0.0_{\pm0.0}$ & $0.0_{\pm0.0}$ & $1.3_{\pm2.3}$ & $65.3_{\pm10.1}$ & $14.7_{\pm8.3}$ \\
\midrule
PerAct & $19.0_{\pm1.5}$ & $0.0_{\pm0.0}$ & $45.3_{\pm8.3}$ & $6.7_{\pm6.1}$ & $6.7_{\pm6.1}$ & $0.0_{\pm0.0}$ & $0.0_{\pm0.0}$ & $20.0_{\pm4.0}$ & $73.3_{\pm8.3}$ \\
\rowcolor{gray!10} \textbf{LexiconVLA (PerAct)} & $19.3_{\pm0.6}$ & $0.0_{\pm0.0}$ & $45.3_{\pm9.2}$ & $4.0_{\pm4.0}$ & $6.7_{\pm4.6}$ & $0.0_{\pm0.0}$ & $0.0_{\pm0.0}$ & $30.7_{\pm4.6}$ & $68.0_{\pm6.9}$ \\
\midrule
RVT & $11.8_{\pm2.3}$ & $0.0_{\pm0.0}$ & $52.0_{\pm8.0}$ & $\mathbf{8.0}_{\pm4.0}$ & $0.0_{\pm0.0}$ & $0.0_{\pm0.0}$ & $0.0_{\pm0.0}$ & $28.0_{\pm4.0}$ & $6.7_{\pm4.6}$ \\
\rowcolor{gray!10} \textbf{LexiconVLA (RVT)} & $14.5_{\pm1.8}$ & $0.0_{\pm0.0}$ & $49.3_{\pm6.1}$ & $5.3_{\pm4.6}$ & $0.0_{\pm0.0}$ & $0.0_{\pm0.0}$ & $0.0_{\pm0.0}$ & $33.3_{\pm12.2}$ & $28.0_{\pm4.0}$ \\
\midrule
RVT-2 & $18.0_{\pm0.5}$ & $0.0_{\pm0.0}$ & $64.0_{\pm4.0}$ & $4.0_{\pm0.0}$ & $1.3_{\pm2.3}$ & $0.0_{\pm0.0}$ & $6.7_{\pm8.3}$ & $42.7_{\pm6.1}$ & $25.3_{\pm4.6}$ \\
\rowcolor{gray!10} \textbf{LexiconVLA (RVT-2)} & $19.8_{\pm0.8}$ & $1.3_{\pm2.3}$ & $73.3_{\pm2.3}$ & $4.0_{\pm4.0}$ & $0.0_{\pm0.0}$ & $0.0_{\pm0.0}$ & $1.3_{\pm2.3}$ & $62.7_{\pm10.1}$ & $16.0_{\pm6.9}$ \\
\midrule
BridgeVLA++$^{*}$ & $19.2_{\pm2.0}$ & $0.0_{\pm0.0}$ & $70.7_{\pm10.1}$ & $5.3_{\pm6.1}$ & $0.0_{\pm0.0}$ & $0.0_{\pm0.0}$ & $2.7_{\pm4.6}$ & $54.7_{\pm8.3}$ & $20.0_{\pm6.9}$ \\
\rowcolor{gray!10} \textbf{LexiconVLA (BridgeVLA++)} & $26.8_{\pm1.2}$ & $22.7_{\pm12.2}$ & $84.0_{\pm0.0}$ & $6.7_{\pm2.3}$ & $0.0_{\pm0.0}$ & $0.0_{\pm0.0}$ & $9.3_{\pm2.3}$ & $62.7_{\pm12.2}$ & $29.3_{\pm8.3}$ \\
\midrule
BridgeVLA & $16.7_{\pm1.3}$ & $9.3_{\pm4.6}$ & $44.0_{\pm0.0}$ & $4.0_{\pm0.0}$ & $2.7_{\pm2.3}$ & $0.0_{\pm0.0}$ & $4.0_{\pm4.0}$ & $57.3_{\pm12.9}$ & $12.0_{\pm4.0}$ \\
\rowcolor{gray!10} \textbf{LexiconVLA (BridgeVLA)} & $\mathbf{34.2}_{\pm2.0}$ & $\mathbf{36.0}_{\pm6.9}$ & $61.3_{\pm2.3}$ & $\mathbf{8.0}_{\pm4.0}$ & $\mathbf{56.0}_{\pm0.0}$ & $0.0_{\pm0.0}$ & $\mathbf{24.0}_{\pm8.0}$ & $\mathbf{82.7}_{\pm8.3}$ & $5.3_{\pm2.3}$ \\
\bottomrule
\end{tabular}}
\end{table*}

\begin{table*}[htbp!]
\centering
\caption{\textbf{Per-task success rate (\%) on the 18 seen tasks.} Mean$_{\pm\text{std}}$ over three random seeds. Best result per column in bold. Shaded rows are LexiconVLA variants. The Avg.\ column covers all 18 tasks.}
\label{tab:sim-seen}
\setlength{\tabcolsep}{4pt}
\renewcommand{\arraystretch}{1.15}
\resizebox{\textwidth}{!}{%
\begin{tabular}{@{}lc|ccccccccc@{}}
\toprule
\multirow{2}{*}{\textbf{Method}} & \multirow{2}{*}{\textbf{Avg.}} & \textbf{Close} & \textbf{Drag} & \textbf{Insert} & \textbf{Meat off} & \textbf{Open} & \textbf{Place} & \textbf{Place} & \textbf{Push} & \textbf{Put in} \\
 &  & \textbf{Jar} & \textbf{Stick} & \textbf{Peg} & \textbf{Grill} & \textbf{Drawer} & \textbf{Cups} & \textbf{Wine} & \textbf{Buttons} & \textbf{Cupboard} \\
\midrule
Act3D & $65.2_{\pm1.2}$ & $\mathbf{100.0}_{\pm0.0}$ & $66.7_{\pm8.3}$ & $13.3_{\pm2.3}$ & $\mathbf{100.0}_{\pm0.0}$ & $90.7_{\pm9.2}$ & $8.0_{\pm4.0}$ & $65.3_{\pm4.6}$ & $\mathbf{100.0}_{\pm0.0}$ & $65.3_{\pm6.1}$ \\
3D Diffuser Actor & $78.3_{\pm1.9}$ & $96.0_{\pm0.0}$ & $98.7_{\pm2.3}$ & $60.0_{\pm8.0}$ & $\mathbf{100.0}_{\pm0.0}$ & $77.3_{\pm12.2}$ & $33.3_{\pm6.1}$ & $90.7_{\pm6.1}$ & $\mathbf{100.0}_{\pm0.0}$ & $81.3_{\pm6.1}$ \\
SAM2Act & $84.3_{\pm1.8}$ & $\mathbf{100.0}_{\pm0.0}$ & $\mathbf{100.0}_{\pm0.0}$ & $80.0_{\pm0.0}$ & $\mathbf{100.0}_{\pm0.0}$ & $90.7_{\pm4.6}$ & $33.3_{\pm4.6}$ & $96.0_{\pm0.0}$ & $\mathbf{100.0}_{\pm0.0}$ & $65.3_{\pm8.3}$ \\
\midrule
PerAct & $37.1_{\pm2.1}$ & $34.7_{\pm4.6}$ & $90.7_{\pm2.3}$ & $0.0_{\pm0.0}$ & $88.0_{\pm4.0}$ & $81.3_{\pm8.3}$ & $0.0_{\pm0.0}$ & $50.7_{\pm15.1}$ & $26.7_{\pm2.3}$ & $6.7_{\pm2.3}$ \\
\rowcolor{gray!10} \textbf{LexiconVLA (PerAct)} & $39.5_{\pm2.1}$ & $9.3_{\pm2.3}$ & $81.3_{\pm4.6}$ & $4.0_{\pm4.0}$ & $78.7_{\pm2.3}$ & $76.0_{\pm10.6}$ & $0.0_{\pm0.0}$ & $50.7_{\pm2.3}$ & $37.3_{\pm2.3}$ & $17.3_{\pm6.1}$ \\
\midrule
RVT & $59.3_{\pm2.0}$ & $36.0_{\pm4.0}$ & $92.0_{\pm0.0}$ & $26.7_{\pm8.3}$ & $97.3_{\pm2.3}$ & $77.3_{\pm6.1}$ & $1.3_{\pm2.3}$ & $92.0_{\pm4.0}$ & $\mathbf{100.0}_{\pm0.0}$ & $36.0_{\pm8.0}$ \\
\rowcolor{gray!10} \textbf{LexiconVLA (RVT)} & $60.8_{\pm1.1}$ & $28.0_{\pm4.0}$ & $94.7_{\pm2.3}$ & $13.3_{\pm2.3}$ & $93.3_{\pm2.3}$ & $82.7_{\pm4.6}$ & $1.3_{\pm2.3}$ & $88.0_{\pm6.9}$ & $\mathbf{100.0}_{\pm0.0}$ & $56.0_{\pm8.0}$ \\
\midrule
RVT-2 & $75.6_{\pm1.0}$ & $18.7_{\pm6.1}$ & $98.7_{\pm2.3}$ & $26.7_{\pm6.1}$ & $\mathbf{100.0}_{\pm0.0}$ & $85.3_{\pm6.1}$ & $50.7_{\pm6.1}$ & $92.0_{\pm8.0}$ & $96.0_{\pm0.0}$ & $57.3_{\pm2.3}$ \\
\rowcolor{gray!10} \textbf{LexiconVLA (RVT-2)} & $76.3_{\pm0.3}$ & $24.0_{\pm4.0}$ & $94.7_{\pm2.3}$ & $44.0_{\pm4.0}$ & $\mathbf{100.0}_{\pm0.0}$ & $88.0_{\pm6.9}$ & $38.7_{\pm6.1}$ & $93.3_{\pm4.6}$ & $96.0_{\pm0.0}$ & $54.7_{\pm6.1}$ \\
\midrule
BridgeVLA++$^{*}$ & $79.7_{\pm1.1}$ & $\mathbf{100.0}_{\pm0.0}$ & $44.0_{\pm6.9}$ & $34.7_{\pm10.1}$ & $\mathbf{100.0}_{\pm0.0}$ & $96.0_{\pm4.0}$ & $\mathbf{56.0}_{\pm4.0}$ & $\mathbf{97.3}_{\pm4.6}$ & $\mathbf{100.0}_{\pm0.0}$ & $81.3_{\pm8.3}$ \\
\rowcolor{gray!10} \textbf{LexiconVLA (BridgeVLA++)} & $77.0_{\pm1.7}$ & $92.0_{\pm0.0}$ & $58.7_{\pm4.6}$ & $36.0_{\pm8.0}$ & $98.7_{\pm2.3}$ & $\mathbf{97.3}_{\pm2.3}$ & $44.0_{\pm4.0}$ & $\mathbf{97.3}_{\pm2.3}$ & $\mathbf{100.0}_{\pm0.0}$ & $76.0_{\pm0.0}$ \\
\midrule
BridgeVLA & $\mathbf{89.8}_{\pm1.0}$ & $\mathbf{100.0}_{\pm0.0}$ & $\mathbf{100.0}_{\pm0.0}$ & $\mathbf{98.7}_{\pm2.3}$ & $\mathbf{100.0}_{\pm0.0}$ & $96.0_{\pm0.0}$ & $45.3_{\pm4.6}$ & $90.7_{\pm6.1}$ & $\mathbf{100.0}_{\pm0.0}$ & $84.0_{\pm4.0}$ \\
\rowcolor{gray!10} \textbf{LexiconVLA (BridgeVLA)} & $87.5_{\pm0.9}$ & $\mathbf{100.0}_{\pm0.0}$ & $\mathbf{100.0}_{\pm0.0}$ & $85.3_{\pm6.1}$ & $\mathbf{100.0}_{\pm0.0}$ & $96.0_{\pm0.0}$ & $54.7_{\pm2.3}$ & $93.3_{\pm8.3}$ & $\mathbf{100.0}_{\pm0.0}$ & $\mathbf{85.3}_{\pm4.6}$ \\
\bottomrule
\end{tabular}}

\vspace{1.2ex}

\resizebox{\textwidth}{!}{%
\begin{tabular}{@{}lccccccccc@{}}
\toprule
\multirow{2}{*}{\textbf{Method}} & \textbf{Put in} & \textbf{Put in} & \textbf{Screw} & \textbf{Slide} & \textbf{Sort} & \textbf{Stack} & \textbf{Stack} & \textbf{Sweep to} & \textbf{Turn} \\
 & \textbf{Drawer} & \textbf{Safe} & \textbf{Bulb} & \textbf{Block} & \textbf{Shape} & \textbf{Blocks} & \textbf{Cups} & \textbf{Dustpan} & \textbf{Tap} \\
\midrule
Act3D & $76.0_{\pm4.0}$ & $\mathbf{100.0}_{\pm0.0}$ & $41.3_{\pm12.9}$ & $96.0_{\pm0.0}$ & $45.3_{\pm2.3}$ & $5.3_{\pm6.1}$ & $12.0_{\pm6.9}$ & $93.3_{\pm6.1}$ & $94.7_{\pm6.1}$ \\
3D Diffuser Actor & $88.0_{\pm0.0}$ & $94.7_{\pm2.3}$ & $82.7_{\pm2.3}$ & $\mathbf{100.0}_{\pm0.0}$ & $12.0_{\pm8.0}$ & $\mathbf{84.0}_{\pm4.0}$ & $40.0_{\pm4.0}$ & $72.0_{\pm4.0}$ & $\mathbf{98.7}_{\pm2.3}$ \\
SAM2Act & $97.3_{\pm4.6}$ & $98.7_{\pm2.3}$ & $94.7_{\pm2.3}$ & $76.0_{\pm4.0}$ & $42.7_{\pm8.3}$ & $73.3_{\pm4.6}$ & $\mathbf{77.3}_{\pm10.1}$ & $98.7_{\pm2.3}$ & $93.3_{\pm2.3}$ \\
\midrule
PerAct & $21.3_{\pm2.3}$ & $34.7_{\pm10.1}$ & $8.0_{\pm0.0}$ & $57.3_{\pm10.1}$ & $0.0_{\pm0.0}$ & $17.3_{\pm2.3}$ & $5.3_{\pm2.3}$ & $58.7_{\pm6.1}$ & $86.7_{\pm6.1}$ \\
\rowcolor{gray!10} \textbf{LexiconVLA (PerAct)} & $58.7_{\pm8.3}$ & $40.0_{\pm6.9}$ & $22.7_{\pm8.3}$ & $61.3_{\pm6.1}$ & $1.3_{\pm2.3}$ & $17.3_{\pm6.1}$ & $1.3_{\pm2.3}$ & $62.7_{\pm2.3}$ & $90.7_{\pm6.1}$ \\
\midrule
RVT & $86.7_{\pm11.5}$ & $65.3_{\pm4.6}$ & $48.0_{\pm6.9}$ & $78.7_{\pm6.1}$ & $33.3_{\pm4.6}$ & $37.3_{\pm9.2}$ & $9.3_{\pm4.6}$ & $58.7_{\pm6.1}$ & $92.0_{\pm0.0}$ \\
\rowcolor{gray!10} \textbf{LexiconVLA (RVT)} & $85.3_{\pm2.3}$ & $64.0_{\pm4.0}$ & $49.3_{\pm6.1}$ & $85.3_{\pm4.6}$ & $33.3_{\pm6.1}$ & $33.3_{\pm2.3}$ & $22.7_{\pm4.6}$ & $74.7_{\pm6.1}$ & $89.3_{\pm2.3}$ \\
\midrule
RVT-2 & $93.3_{\pm6.1}$ & $\mathbf{100.0}_{\pm0.0}$ & $\mathbf{97.3}_{\pm2.3}$ & $70.7_{\pm6.1}$ & $46.7_{\pm4.6}$ & $76.0_{\pm4.0}$ & $61.3_{\pm4.6}$ & $96.0_{\pm0.0}$ & $94.7_{\pm2.3}$ \\
\rowcolor{gray!10} \textbf{LexiconVLA (RVT-2)} & $\mathbf{98.7}_{\pm2.3}$ & $98.7_{\pm2.3}$ & $94.7_{\pm4.6}$ & $72.0_{\pm6.9}$ & $42.7_{\pm4.6}$ & $73.3_{\pm6.1}$ & $72.0_{\pm4.0}$ & $\mathbf{100.0}_{\pm0.0}$ & $88.0_{\pm4.0}$ \\
\midrule
BridgeVLA++$^{*}$ & $93.3_{\pm4.6}$ & $96.0_{\pm4.0}$ & $85.3_{\pm2.3}$ & $94.7_{\pm6.1}$ & $58.7_{\pm4.6}$ & $68.0_{\pm6.9}$ & $37.3_{\pm11.5}$ & $98.7_{\pm2.3}$ & $93.3_{\pm2.3}$ \\
\rowcolor{gray!10} \textbf{LexiconVLA (BridgeVLA++)} & $96.0_{\pm0.0}$ & $97.3_{\pm2.3}$ & $80.0_{\pm4.0}$ & $\mathbf{100.0}_{\pm0.0}$ & $61.3_{\pm6.1}$ & $70.7_{\pm6.1}$ & $12.0_{\pm6.9}$ & $76.0_{\pm4.0}$ & $93.3_{\pm4.6}$ \\
\midrule
BridgeVLA & $97.3_{\pm2.3}$ & $98.7_{\pm2.3}$ & $\mathbf{97.3}_{\pm2.3}$ & $97.3_{\pm4.6}$ & $\mathbf{70.7}_{\pm2.3}$ & $77.3_{\pm8.3}$ & $70.7_{\pm8.3}$ & $\mathbf{100.0}_{\pm0.0}$ & $92.0_{\pm0.0}$ \\
\rowcolor{gray!10} \textbf{LexiconVLA (BridgeVLA)} & $96.0_{\pm6.9}$ & $\mathbf{100.0}_{\pm0.0}$ & $92.0_{\pm0.0}$ & $92.0_{\pm0.0}$ & $57.3_{\pm8.3}$ & $76.0_{\pm6.9}$ & $56.0_{\pm6.9}$ & $\mathbf{100.0}_{\pm0.0}$ & $90.7_{\pm4.6}$ \\
\bottomrule
\end{tabular}}
\end{table*}